\documentclass[10pt,twocolumn,letterpaper]{article}

\usepackage[pagenumbers]{cvpr}
\usepackage{fix-cm}

\usepackage{microtype}
\usepackage{tikz}
\usepackage{booktabs}
\usepackage{colortbl}
\usepackage{multirow}
\usepackage{array}
\usepackage{tabularx}
\usepackage{xspace}
\usepackage{pifont}
\usepackage{fontawesome5}
\providecommand{\Checkmark}{\ding{51}}
\providecommand{\XSolidBrush}{\ding{55}}
\definecolor{tablecheck}{rgb}{0,0.65,0.31}
\definecolor{tablecross}{rgb}{0.91,0.59,0.48}
\newcommand{\cmark}{\textcolor{tablecheck}{\Checkmark}}
\newcommand{\xmark}{\textcolor{tablecross}{\XSolidBrush}}
\definecolor{tablerowalt}{gray}{0.95}
\definecolor{tablehead}{HTML}{E0E0E0}
\definecolor{thinkwamrow}{HTML}{E1F5FE}
\definecolor{thinkwamline}{RGB}{0,126,102}
\definecolor{boxfill}{RGB}{246,249,250}
\renewcommand{\arraystretch}{1.12}
\newcommand{\thinkwam}{ThinkWAM\xspace}

\makeatletter
\patchcmd{\@maketitle}{\vskip .375in}{\vskip 0pt}{}{}
\patchcmd{\@maketitle}{\vspace*{24pt}}{\vspace*{10pt}}{}{}
\patchcmd{\@maketitle}{\vspace*{12pt}}{\vspace*{4pt}}{}{}
\makeatother

\usepackage{float}
\usepackage{stfloats} 
\usepackage{placeins} 
\usepackage{algorithm}
\usepackage{algpseudocode}
\renewcommand{\dbltopfraction}{0.92}
\renewcommand{\topfraction}{0.95}

\renewcommand{\dblfloatpagefraction}{0.8}
\renewcommand{\textfraction}{0.05}

\usepackage{tcolorbox}
\definecolor{thinktitle}{HTML}{244F73}
\definecolor{thinkpaperbg}{HTML}{F0F7FB}
\makeatletter
\newcommand{\thinkauthors}{\@author}
\makeatother

\definecolor{cvprblue}{rgb}{0.21,0.49,0.74}
\usepackage[breaklinks,colorlinks,allcolors=cvprblue]{hyperref}

\def\confName{CVPR}
\def\confYear{2027}

\title{ThinkWAM: Rethinking Latent Futures for Action Refinement}

\author{%
\fontsize{9.3}{11}\selectfont Haoyun Tang$^{1,*}$\quad Baqiao Deng$^{1,*}$\quad Enjin Zhou$^{1}$\quad Songwen Tian$^{1}$\quad Keyao Xu$^{2}$\\[2pt]
\fontsize{9.3}{11}\selectfont Bowen Jing$^{3}$\quad Kun Wang$^{4}$\quad Ke Ma$^{1,5}$\quad Zhandong Mei$^{1,\dagger}$\quad Huibin Li$^{1,5,\dagger}$\\[4pt]
\fontsize{8.2}{10}\selectfont $^{1}$Xi'an Jiaotong University\quad $^{2}$Nankai University\quad $^{3}$The University of Manchester\\[1pt]
\fontsize{8.2}{10}\selectfont $^{4}$Nanyang Technological University\qquad $^{5}$Phantrix Research\\[2pt]
\fontsize{8}{9}\selectfont $^{*}$Equal contribution.\qquad $^{\dagger}$Co-corresponding authors.
}
\date{}
\hypersetup{
    pdftitle={ThinkWAM: Rethinking Latent Futures for Action Refinement},
    pdfauthor={Haoyun Tang, Baqiao Deng, Enjin Zhou, Songwen Tian, Keyao Xu, Bowen Jing, Kun Wang, Ke Ma, Zhandong Mei, Huibin Li},
    pdfsubject={World-action models for robot manipulation},
    pdfkeywords={World-action models, action refinement, affordance-guided memory, multi-horizon latent prediction}
}

\begin{document}
\makeatletter
\g@addto@macro\normalsize{%
  \setlength{\abovedisplayskip}{6pt plus 2pt minus 2pt}%
  \setlength{\belowdisplayskip}{6pt plus 2pt minus 2pt}%
  \setlength{\abovedisplayshortskip}{3pt plus 2pt}%
  \setlength{\belowdisplayshortskip}{4pt plus 2pt minus 2pt}}
\makeatother
\normalsize
\onecolumn
\thispagestyle{plain}
\vspace*{-40pt}
\pdfbookmark[0]{ThinkWAM}{thinkwam-title}
\begin{tcolorbox}[width=\textwidth,colback=thinkpaperbg,colframe=thinkpaperbg,
  boxrule=0pt,arc=5pt,left=10pt,right=10pt,top=8pt,bottom=6pt,
  before skip=0pt,after skip=6pt]
\begin{center}
{\sffamily\bfseries\color{thinktitle}\fontsize{17}{19}\selectfont
ThinkWAM: Rethinking Latent Futures\\for Action Refinement\par}
\vspace{5pt}
\begin{tabular}{@{}c@{}}\thinkauthors\end{tabular}
\end{center}
\vspace{-3pt}
\begingroup
\fontsize{10}{12}\selectfont
\renewenvironment{abstract}{\normalfont\noindent}{\par}
\begin{abstract}
World-action models (WAMs) couple action generation with predictions of future states. However, an initial action chunk may deviate from the correct plan implied by the predicted future, and the predicted future often fails to reflect how it relates to the specific action output, leaving the prediction--action coupling underexploited and making the action hard to correct before execution. We introduce \textbf{\thinkwam}, a latent WAM that establishes bidirectional feedback before execution through a \emph{proposal--rethink--refine} loop: the policy first proposes a complete action chunk, the proposal corrects multi-horizon future latents through a learned residual, and the corrected latents are injected into a second pass of the same action head, which generates the executable chunk. An affordance-guided observation-action memory and a near-fine, far-coarse multi-horizon prediction scheme, both motivated by predictive coding, support this loop, and all prediction stays in latent space. \thinkwam{} achieves 98.70\% success on LIBERO, 84.76\% on LIBERO-Plus without OOD fine-tuning, 85.75\%/65.85\% on RoboTwin 2.0 Clean/Random, and 65.17\% on RoboCasa24. \thinkwam{} achieves $64.4\%$ mean success across three real-world OOD conditions. The mean two-pass inference latency per action chunk is 153.48\,ms on LIBERO, 164.61\,ms on RoboTwin, and 101.2\,ms on the Galaxea A1Z. Ablations show complementary gains from rethinking and affordance guidance. Diagnostics indicate that proposal-conditioned corrections change both predicted futures and action trajectories, consistent with the gains from rethinking in these ablations.

\end{abstract}

\endgroup
\vspace{4pt}
\noindent
\begin{minipage}{\linewidth}
\centering
\includegraphics[width=0.94\linewidth]{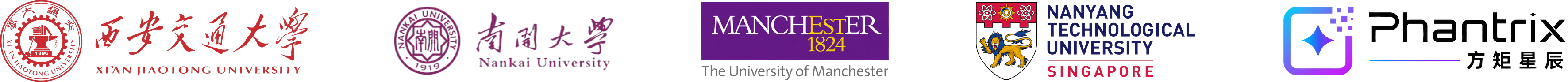}\par
\vspace{4pt}
{\fontsize{9}{11}\selectfont
\href{https://thinkwam-project.github.io/}{\raisebox{-0.4pt}{\faGlobe}\hspace{4pt}\textsf{\textbf{Project page}}}\hspace{18pt}%
\href{https://github.com/HaoyunT/Thinkwam}{\raisebox{-1.3pt}{\includegraphics[height=9pt]{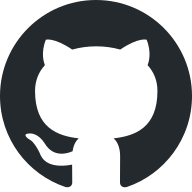}}\hspace{4pt}\textsf{\textbf{Code}}}\hspace{18pt}%
\href{https://huggingface.co/Adrain11/ThinkWAM}{\raisebox{-1.3pt}{\includegraphics[height=9pt]{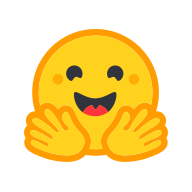}}\hspace{4pt}\textsf{\textbf{Hugging Face}}}\par}
\end{minipage}
\end{tcolorbox}
\begingroup\centering
\includegraphics[width=0.94\textwidth]{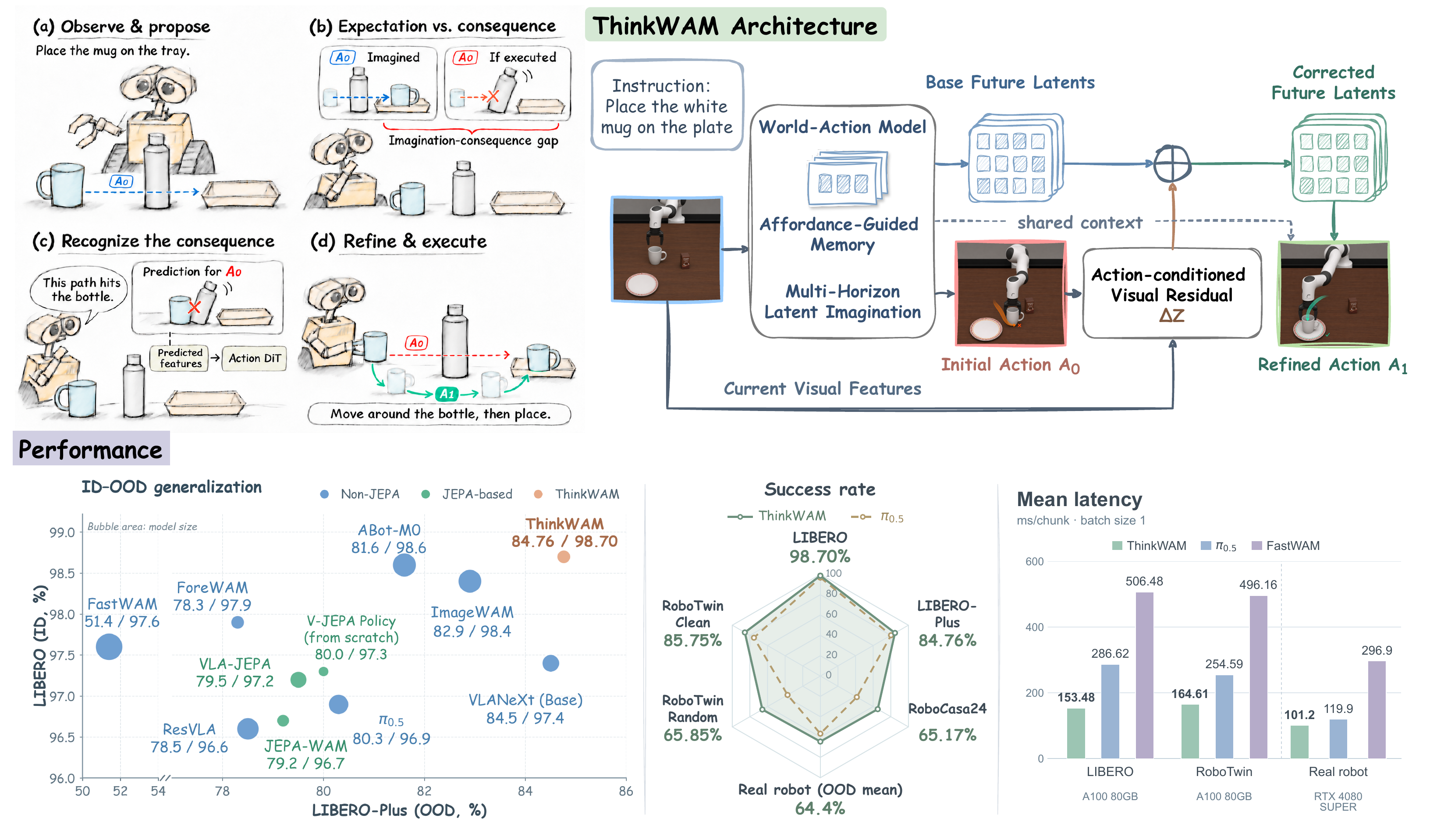}
\captionsetup{font=small,labelfont=bf,skip=4pt,hypcap=false}
\captionof{figure}{Overview of ThinkWAM. The policy first proposes a complete action chunk $A^{(0)}$, then rethinks by correcting its predicted near, mid, and far future latents with this proposal, and finally re-predicts the executable chunk $A^{(1)}$ with the same ActionDiT. The lower panel summarizes success rates and mean inference latency.}
\label{fig:teaser}
\par\endgroup

\clearpage
\twocolumn
\raggedbottom
\makeatletter
\let\thinkwamSavedParagraph\paragraph
\renewcommand\paragraph{\@startsection{paragraph}{4}{\z@}%
  {1.5ex plus 0.4ex minus 0.2ex}{-1em}%
  {\normalfont\normalsize\bfseries}}
\makeatother
\begingroup
\linespread{1.0}\selectfont
\section{Introduction}
\label{sec:intro}

Robot manipulation requires anticipating what an action will cause. A visually plausible reach can miss an object, an apparently stable grasp can slip, and a locally reasonable motion can leave a later placement incomplete. Vision-language-action (VLA) models connect visual observations and language instructions to robot control, and scalable robot transformers, web-supervised visual-language backbones, and generalist policies have broadened the range of tasks that a single policy can perform~\cite{brohan2022rt1,brohan2023rt2visionlanguageactionmodelstransfer,octomodelteam2024octoopensourcegeneralistrobot,kim2024openvlaopensourcevisionlanguageactionmodel,pi05}. Diffusion and flow-matching action heads further support continuous, multimodal action chunks~\cite{chi2023diffusionpolicy,lipman2023flowmatching,black2024pi0}. Yet the usual decision interface maps the current visual context directly to an action or action chunk, without an explicit representation of what it will cause.

World-action models (WAMs) address this gap by coupling action generation with a predictive representation of the environment~\cite{du2023learninguniversalpoliciestextguided,wu2023unleashinglargescalevideogenerative,cen2025worldvlaautoregressiveactionworld}, and latent-space variants avoid reconstructing every future pixel~\cite{hafner2019learninglatentdynamicsplanningpixels,assran2025vjepa2,lin2026jepawam,zhang2026vjepapolicy,huang2026foresight}. These models use future prediction as training supervision, expose future-informed features to the action head, or use predicted consequences to guide action sampling~\cite{an2026feedbackwm}. Here, we study how a complete action proposal can update predicted futures and how those updates can guide action refinement before execution (Sec.~\ref{sec:related-latent}).

An initial action chunk may be flawed. If future prediction is independent of that chunk, a mistimed grasp and a well-timed grasp can receive the same predicted continuation, limiting its usefulness for refinement. We therefore seek feedback in both directions before execution: predicted futures guide the action, while the proposal updates those futures. We introduce \thinkwam{}, a latent WAM that implements this interaction through a \emph{proposal--rethink--refine} loop (Fig.~\ref{fig:teaser}). The policy predicts visual latents at multiple future horizons and uses language features and affordance-guided memory to propose a complete action chunk. The proposal then conditions learned residual corrections to the base future latents. A second pass through the \emph{same} ActionDiT head consumes the corrected futures and generates the refined action chunk. Sensory--motor accounts motivate this loop: agents condition sensory predictions on their own motor commands to guide behavior~\cite{rao2024sensory}. 

Two ideas from predictive coding motivate our representation choices. Predictive coding shapes representations through prediction-error minimization~\cite{rao1999predictive}. We use affordance supervision to shape the shared visual context toward where actions make contact, and write a weighted summary of this context into memory. Dynamic predictive coding motivates hierarchical prediction across temporal scales~\cite{jiang2024dynamic}. We therefore allocate more tokens and use finer aggregation for near-future targets than for distant ones.

Our contributions are:
\begin{itemize}
    \item We introduce \thinkwam{}, a proposal--rethink--refine architecture that corrects multi-horizon future latents using the policy's own proposal and injects the corrected latents into a second pass of the same action head, which generates the final action.
    \item We propose an affordance-guided observation-action memory and a near-fine, far-coarse multi-horizon prediction scheme, motivated by predictive coding.
    \item We demonstrate strong generalization on LIBERO-Plus, the randomized setting of RoboTwin 2.0, RoboCasa24, and real-world manipulation tasks, with \thinkwam{} achieving competitive overall results while maintaining low inference latency.
\end{itemize}

\section{Related Work}
\label{sec:related}

\paragraph{VLA policies and world-action models.}
\label{sec:related-vla}
Visuomotor and VLA policies map the current context directly to the next action~\cite{brohan2022rt1,brohan2023rt2visionlanguageactionmodelstransfer,kim2024openvlaopensourcevisionlanguageactionmodel} or to an action chunk produced by chunked, diffusion, or flow-matching heads~\cite{octomodelteam2024octoopensourcegeneralistrobot,pi05,zhao2023learningfinegrainedbimanualmanipulation,chi2023diffusionpolicy,lipman2023flowmatching}, without modeling what these actions will lead to.

World-action models (WAMs) make future states explicit alongside actions. Video-conditioned policies use predicted future frames to guide control~\cite{du2023learninguniversalpoliciestextguided}, whereas other models predict future frames and actions jointly, from a GPT-style causal Transformer that is pretrained on large-scale video for future-frame generation and then fine-tuned to output actions and future images together~\cite{wu2023unleashinglargescalevideogenerative}, to autoregressive action--image modeling~\cite{cen2025worldvlaautoregressiveactionworld}, coupled video--action diffusion~\cite{zhu2025uwm,bi2025motus}, and action encoding as latent frames of a pretrained video model~\cite{kim2026cosmospolicy}. Because generating future video is costly, FastWAM and ImageWAM ask whether test-time imagination or full video generation is necessary~\cite{yuan2026fastwam,imagewam}. Cycle WAM enforces video--action consistency with a training-time cycle-reconstruction objective~\cite{cyclewam2026}, whereas \thinkwam{} lets the policy's own action proposal correct the predicted future at inference (Sec.~\ref{sec:rethink}).

Another line replaces raw appearance with physically grounded cues closer to what an action must change, such as point trajectories~\cite{wen2023atm}, dynamic-region, spatial, and semantic knowledge~\cite{zhang2025dreamvla}, affordance~\cite{mo2021where2actpixelsactionsarticulated}, and affordance and depth guidance for video and action features~\cite{zhang2026devadecoupledvideoactionmodel}. \thinkwam{} follows this direction by injecting an affordance-weighted residual into the memory that conditions both action passes.

\paragraph{Latent world-action models.}
\label{sec:related-latent}
Latent WAMs predict future states in a learned representation space rather than in pixels, commonly that of joint-embedding predictive architectures (JEPAs)~\cite{assran2023selfsupervisedlearningimagesjointembedding,bardes2024revisitingfeaturepredictionlearning,assran2025vjepa2}, which learn by predicting the representations of masked targets. They differ mainly in how future predictions inform action selection.

One route uses future prediction as auxiliary supervision that shapes the policy's representations: JEPA-WAM learns a joint current--future target in a V-JEPA space and drops the transition head at deployment~\cite{lin2026jepawam}, and FRAPPE aligns future representations with multiple visual teachers during training and retains its future-prefix streams at inference~\cite{zhao2026frappe}. A second route supplies future-predictive features to the action expert at inference. V-JEPA Policy exposes the predictor's context-position keys and values (K/V) to its action expert~\cite{zhang2026vjepapolicy}; ForeWAM reads hidden K/V at future positions~\cite{huang2026foresight}; LiLa-WAM jointly produces action velocities and future-related representations in a shared stream~\cite{lilawam2026}. LaWAM decodes a policy-predicted latent action into a future feature that conditions its action expert~\cite{lawam2026}.

\paragraph{Action review and refinement.}
Several methods use predictive feedback to review or revise action proposals during control. Staircase Policy re-predicts future latents at sub-chunk boundaries to update unexecuted actions~\cite{staircase2026}. Feedback World Model predicts action-conditioned outcomes and corrects its state estimate from observed outcomes to guide diffusion-policy sampling~\cite{an2026feedbackwm}. PACT-WAM's Proposal Review decodes predicted images to approve an execution prefix or reject and resample a joint proposal~\cite{liu2026pactwam}. $\tau_0$-WM ranks sampled candidates by re-denoising consistency and, when the best is unreliable, simulates the candidates' futures, scores their task progress, and regenerates an action conditioned on the most promising future~\cite{zhou2026tau0wm}. ThinkWAM instead predicts a proposal-conditioned residual on multi-horizon base future latents and injects the corrected latents into a second pass of the same action head (Sec.~\ref{sec:rethink}); the path is trained jointly, with demonstration futures supervising the predicted latents and the refined-action loss shaping the corrected representation.

\paragraph{Multi-horizon prediction.}
\label{sec:related-horizon}
Predicting at several temporal distances trades off local precision against longer-range foresight. Prior work varies the horizon of predicted actions~\cite{moh2025} or predicts future representations at multiple horizons~\cite{ding2026hifor,nativeva2026}. Some methods also treat near and far predictions differently: some vary noise levels with temporal distance~\cite{ruhe2024rolling,chen2024diffusionforcing}, HiFOR adds state supervision only to its short-term horizon~\cite{ding2026hifor}, Coarse-to-Control plans far ahead with coarse action tokens~\cite{wu2026coarse2control}, and PACT-WAM encodes older history more coarsely~\cite{liu2026pactwam}. Dynamic predictive coding provides a broader motivation for hierarchical, multi-timescale prediction~\cite{jiang2024dynamic}. These treatments vary noise, supervision, or granularity with temporal distance, but mostly for generation, planning, or history encoding. As a model-specific design, \thinkwam{} applies a near-fine, far-coarse allocation to its future latent targets, assigning more query tokens to near horizons. This retains near-term detail and reduces the burden of predicting uncertain distant features. The corrected futures are used at inference rather than only \mbox{as auxiliary supervision.}\looseness=1

\par\endgroup
\section{Method}
\label{sec:method}

\subsection{Overview}
\label{sec:overview}
\begin{figure*}[t]
    \centering
    \includegraphics[width=\textwidth,trim=11bp 0bp 5bp 0bp,clip]{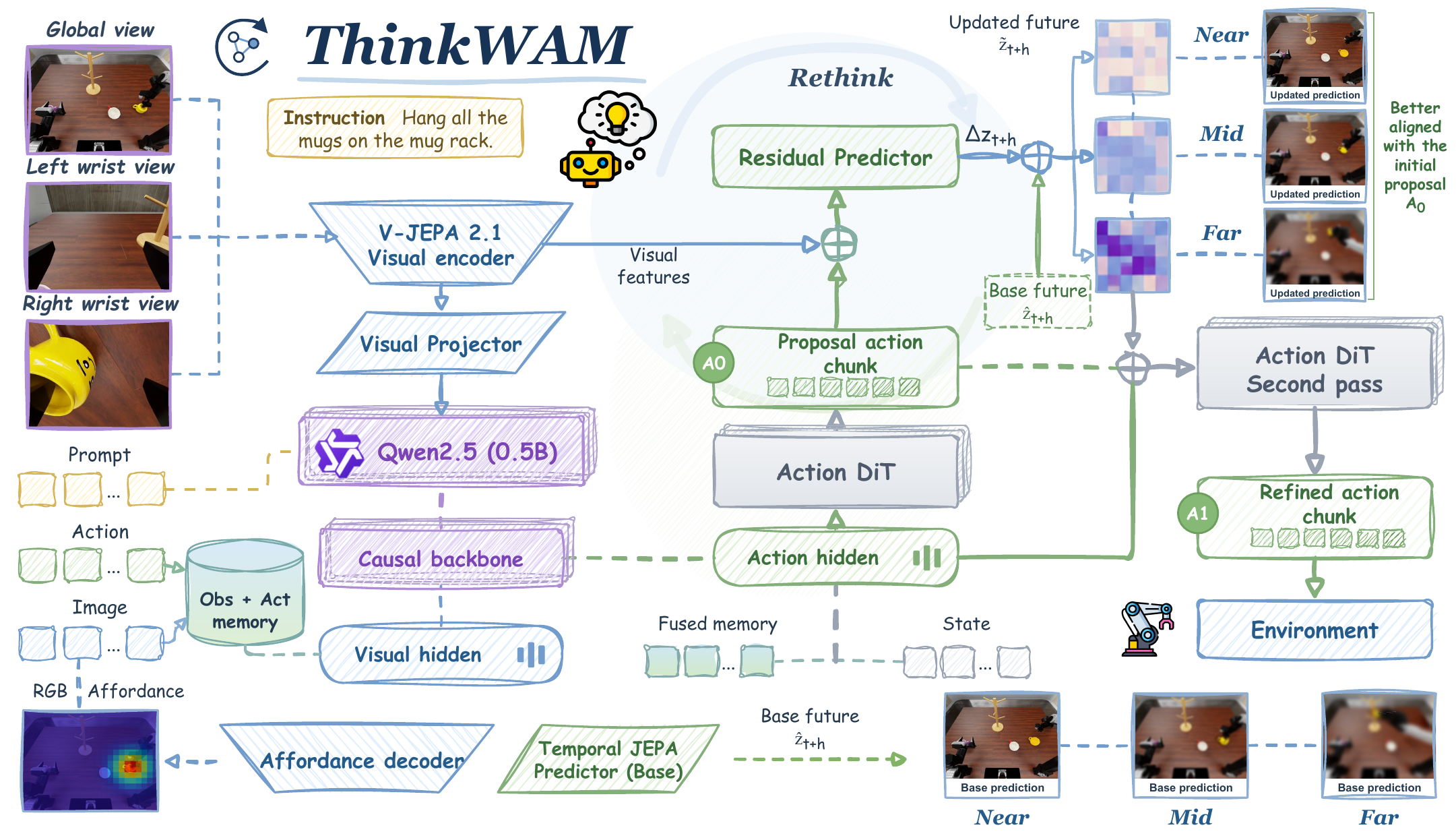}
    \captionsetup{width=\textwidth}
    \caption{Architecture of ThinkWAM. The temporal JEPA predicts base future latents from the visual prefix and memory, and the affordance decoder adds a memory residual. The proposal $A_t^{(0)}$ from the first ActionDiT pass conditions the residual predictor, and the corrected futures condition the second pass, which outputs $A_t^{(1)}$.}
    \label{fig:overview}
\end{figure*}

ThinkWAM builds an affordance-guided observation-action memory (Sec.~\ref{sec:memory}) and near-fine, far-coarse base future latents (Sec.~\ref{sec:imagination}), then runs a proposal--rethink--refine loop through a shared action head (Sec.~\ref{sec:rethink}, Fig.~\ref{fig:overview}): a first pass proposes a complete chunk, the proposal corrects the futures through a latent residual, and a second pass generates the executable chunk from the corrected futures.

\subsection{Problem Setup}
\label{sec:problem}
At decision time $t$, the policy receives multi-view visual observations $o_t$, an instruction $\ell$, proprioception $q_t$, and the $L$ most recently executed actions $\mathcal{A}_t=\{a_{t-L},\ldots,a_{t-1}\}$. It predicts an action chunk $A_t=(a_t,\ldots,a_{t+K-1})$. Alongside the chunk, ThinkWAM predicts future visual latents at horizons $\mathcal{H}$, with $N_h$ tokens per horizon $h$.

ThinkWAM builds on three components. A frozen V-JEPA 2.1 ViT-L encoder~\cite{murlabadia2026vjepa21} $E_v$ maps observations to visual tokens $z_t=E_v(o_t)$ and also defines the future targets. A Qwen2.5-0.5B language backbone~\cite{qwen2025qwen25} receives projected visual tokens and $\ell$, producing visual-prefix hidden states $H_t^v$ and language-action features $f_t^{\mathrm{lang\text{-}act}}$. A diffusion-Transformer action head, ActionDiT, generates a chunk from a conditioning context $c$ and $q_t$ by flow matching~\cite{lipman2023flowmatching}: with $x_\tau=(1-\tau)\epsilon+\tau A^*$, Gaussian $\epsilon$, and a sampled time $\tau$, its velocity $v_\theta$ is trained with
\begin{equation}
    \mathcal{L}_{\mathrm{FM}}(c,A^*)=\mathbb{E}_{\tau,\epsilon}\left\|v_\theta(x_\tau,\tau;c,q_t)-(A^*-\epsilon)\right\|_2^2,
\end{equation}
where $A^*$ is the expert chunk.

\subsection{Affordance-Guided Memory}
\label{sec:memory}
\label{sec:affordance}
\paragraph{Memory and temporal context.}
We first form a memory representation from the current visual tokens and recently executed actions:
\begin{equation}
    m_t = \mathcal{M}(z_t,\mathcal{A}_t) \in \mathbb{R}^{S\times d_m},
\end{equation}
where $S$ is the number of learned memory slots. Each slot cross-attends to pooled visual tokens and projected action-history tokens and is then updated by a feed-forward layer. The memory is a compact summary of the current observation and recent actions: it is recomputed at every decision step, and past information enters only through the executed actions. A temporal context transformer then encodes the projected visual prefix together with the memory slots,
\begin{equation}
    U_t=\mathcal{T}\left([\phi_v(H_t^v);\phi_m(m_t)]\right),
\end{equation}
where $\phi_v$, $\phi_m$ are learned projections, so that $U_t=[U_t^v;U_t^m]$ consists of a visual prefix $U_t^v$ and a memory suffix $U_t^m$. This shared context serves both the affordance decoder below and the future predictor (Sec.~\ref{sec:imagination}).

\paragraph{Affordance guidance.}
Following the predictive-coding view that a representation is shaped by what it must predict~\cite{rao1999predictive}, ThinkWAM trains the shared context to support predicting where the action will make contact. An affordance decoder reads the visual prefix $U_t^v$ and predicts one affordance probability $p_{t,n}$ per visual token. It is supervised with $\mathcal{L}_{\mathrm{aff}}=\mathrm{BCE}+\lambda_{\mathrm{Dice}}\,\mathrm{Dice}$ against contact heatmaps that are area-averaged onto each view's token grid as soft labels. Each visual-prefix token is then weighted by the decoder's certainty $\rho_{t,n}=2|p_{t,n}-\tfrac12|$ to form a summary that is written into the memory:
\begin{equation}
    s_t^{\mathrm{aff}} = \frac{\sum_n \rho_{t,n}U^v_{t,n}}{\sum_n \rho_{t,n}+\delta},
    \qquad
    \Delta m_t^{\mathrm{aff}} = g_{\mathrm{aff}}(s_t^{\mathrm{aff}}),
\end{equation}
where $\delta$ is a small constant. The memory residual $\Delta m_t^{\mathrm{aff}}$ is added to all memory slots, giving the guided memory
\begin{equation}
    m_t^{\mathrm{guide}} = m_t + \alpha_t\Delta m_t^{\mathrm{aff}},
    \label{eq:guided-memory}
\end{equation}
where $\alpha_t$ combines a training schedule with a gate on the mean certainty (Appendix~\ref{app:affordance-implementation}). Because $\rho_{t,n}$ measures certainty rather than affordance probability, the summary is not restricted to affordance regions. Affordance thus influences the actions in two ways: during training, its supervision passes gradients into the shared context, and the residual updates the memory at each decision step. The future predictor reads the base memory $m_t$, so only the two action passes read $m_t^{\mathrm{guide}}$.

\subsection{Near-Fine, Far-Coarse Prediction}
\label{sec:imagination}
Manipulation couples local geometric changes with longer-term state transitions. Approaching an object and adjusting a grasp require nearby visual detail, whereas transport and subsequent placement span a larger part of the demonstration. Dynamic predictive coding motivates hierarchical prediction across temporal scales~\cite{jiang2024dynamic}. We therefore use a near-fine, far-coarse token allocation for future targets.

In the frozen V-JEPA feature space, a temporal JEPA predictor reads $U_t$ via horizon-specific learned queries:
\begin{equation}
    \hat z_{t+h}=P_h\left(Q_h,\;U_t\right),
    \qquad h\in\mathcal{H}.
\end{equation}
Here $Q_h$ is the set of $N_h$ learned query tokens of horizon $h$, each of which yields one token of $\hat z_{t+h}$, and $P_h$ is the horizon-specific head. All heads predict directly from the current context, without recursively consuming another head's prediction. Each target pools the future V-JEPA features to $N_h$ tokens:
\begin{equation}
    y_{t+h}=\operatorname{sg}\!\left[\operatorname{Norm}\!\left(
    \operatorname{Pool}_{N_h}(E_v(o_{t+h}))\right)\right],
    \label{eq:future-target}
\end{equation}
where $\operatorname{Pool}_{N_h}$ is adaptive average pooling over the flattened token sequence, and $\operatorname{sg}$ stops target gradients. Each token retains the same feature dimension. The budget $N_h$ decreases with $h$: on LIBERO, $\mathcal{H}=\{7,15,31\}$ with $(N_7,N_{15},N_{31})=(288,144,72)$. More tokens preserve finer feature detail at near horizons, while stronger aggregation at distant horizons reduces the demand to predict uncertain local detail. The predictor is trained with the mean token-wise cosine distance,
\begin{equation}
\begin{aligned}
    \mathcal{L}_h &= \frac{1}{B N_h}\sum_{b=1}^{B}\sum_{i=1}^{N_h}
    \left[1-\cos(\hat z_{t+h,b,i},y_{t+h,b,i})\right],\\
    \mathcal{L}_{\mathrm{jepa}} &= \frac{1}{|\mathcal{H}|}\sum_{h\in\mathcal{H}}\mathcal{L}_h,
\end{aligned}
\end{equation}
where $B$ is the batch size, so that all horizons are weighted equally despite their different token counts. Because this loss also updates the memory and backbone shared with action generation, these representations must retain information that predicts later observations. Since the visual prefix precedes the text under causal attention, the base prediction does not explicitly depend on the instruction and is identical for every candidate chunk; only the latent residual described next makes the futures depend on the proposal.


\subsection{Proposal--Rethink--Refine}
\label{sec:rethink}
\paragraph{Proposal.}
The language-action features and the guided memory, with the latter projected by a learned map $W_m$, form the first action context, from which the first ActionDiT pass predicts a complete initial chunk:
\begin{equation}
\begin{aligned}
    c_t^{(0)}&=[f_t^{\mathrm{lang\text{-}act}};W_m m_t^{\mathrm{guide}}],\\
    A_t^{(0)}&=\operatorname{ActionDiT}(c_t^{(0)},q_t).
\end{aligned}
\end{equation}
The complete chunk conditions the latent residual model.

\paragraph{Rethink.}
The base prediction $\hat z_{t+h}$ describes a demonstration continuation inferred from current evidence and executed actions, whereas $A_t^{(0)}$ specifies the control sequence the policy intends to execute. Motivated by sensory--motor accounts in which an agent conditions sensory predictions on its own actions to guide behavior~\cite{rao2024sensory}, we let the proposal correct the predicted futures before any action of the new chunk is executed. A compact latent residual model reads the proposal together with the current visual tokens:
\begin{equation}
\begin{aligned}
    \Delta z_{t+h}&=R_h(z_t,A_t^{(0)}),\\
    \tilde z_{t+h}&=\operatorname{sg}[\hat z_{t+h}]+\Delta z_{t+h},
    \qquad h\in\mathcal{H}.
\end{aligned}
    \label{eq:corrected-latent}
\end{equation}
The base prediction acts as a visual reference; the stop-gradient keeps its forward value but blocks direct updates from the refinement path, so $\hat z_{t+h}$ remains trained by $\mathcal{L}_{\mathrm{jepa}}$.
The residual is trained to predict the difference between the observed future features and the base prediction:
\begin{equation}
\begin{aligned}
    r_h^*&=\operatorname{sg}\big[\operatorname{Pool}_{N_h}(E_v(o_{t+h}))-\hat z_{t+h}\big],\\
    \mathcal{L}_{\mathrm{res}}&=\frac{1}{|\mathcal{H}|}\sum_{h\in\mathcal{H}}\operatorname{SmoothL1}\left(\Delta z_{t+h},\,r_h^*\right).
\end{aligned}
\end{equation}
Since $r_h^*$ is defined by demonstrated futures, $\tilde z_{t+h}$ is not a predicted outcome of $A_t^{(0)}$; rather, $\mathcal{L}_{\mathrm{res}}$ keeps the adjustment visually anchored toward the expert future, while the refined-action loss trains the residual to bridge the gap between the base future prediction and the action proposal, mapping $A_t^{(0)}$ onto the re-corrected visual representation that conditions the second pass. Appendix~\ref{app:multihorizon-implementation} further discusses this design and why we do not supervise the residual with actual rollout outcomes.

\paragraph{Refine.}
Each corrected latent is normalized, mapped to the action-context dimension by a horizon-specific map $W_h$, and concatenated across horizons. The second pass reuses the ActionDiT weights of the first, and its context extends $c_t^{(0)}$ with these corrected future tokens:
\begin{equation}
\begin{aligned}
    c_t^{\mathrm{future}}&=\operatorname{Concat}_{h\in\mathcal{H}}W_h\operatorname{Norm}(\tilde z_{t+h}),\\
    c_t^{(1)}&=[c_t^{(0)};c_t^{\mathrm{future}}],\\
    A_t^{(1)}&=\operatorname{ActionDiT}(c_t^{(1)},q_t).
\end{aligned}
    \label{eq:refined-action}
\end{equation}

\subsection{Training and Inference}
\label{sec:objective}
ThinkWAM is trained end-to-end with
\begin{equation}
\begin{split}
\mathcal{L}={}&\lambda_0\mathcal{L}_{\mathrm{FM}}(c_t^{(0)},A^*)+
\lambda_1\mathcal{L}_{\mathrm{FM}}(c_t^{(1)},A^*)\\
&+\lambda_{\mathrm{jepa}}\mathcal{L}_{\mathrm{jepa}}+
\lambda_{\mathrm{res}}\mathcal{L}_{\mathrm{res}}+
\lambda_{\mathrm{aff}}\mathcal{L}_{\mathrm{aff}},
\end{split}
\end{equation}
which combines the two action passes and the latent residual (Sec.~\ref{sec:rethink}), the affordance decoder (Sec.~\ref{sec:memory}), and the multi-horizon prediction (Sec.~\ref{sec:imagination}). Both passes are supervised against the same expert chunk, with the second-pass loss also shaping the corrected future representations to support action generation. During training, the proposal that conditions the residual model is obtained by differentiably integrating the shared action head from Gaussian noise.

At inference, the future targets and affordance labels are unavailable and unused. The policy forms the guided memory and the base futures, integrates the first pass from Gaussian noise to obtain $A_t^{(0)}$, corrects the futures, injects the corrected latents into the context of the second pass, and integrates that pass, again from freshly sampled Gaussian noise, to obtain $A_t^{(1)}$, which is the only chunk executed.

\par

\section{Experiments}
\label{sec:experiments}

\begin{table*}[!t]
\centering
\caption{LIBERO success (\%) across four suites, with model size in billions (B) and action-supervised policy pretraining (Action P.T.).}
\label{tab:libero}
\small
\renewcommand{\arraystretch}{0.90}
\setlength{\aboverulesep}{0.5pt}
\setlength{\belowrulesep}{0.75pt}
\setlength{\tabcolsep}{4pt}
\renewcommand{\cmark}{\textcolor{tablecheck}{\scalebox{0.9}{\Checkmark}}}
\renewcommand{\xmark}{\textcolor{tablecross}{\scalebox{0.9}{\XSolidBrush}}}
\begin{tabularx}{\textwidth}{lcc*{5}{>{\raggedleft\arraybackslash}X}}
\toprule
\rowcolor{tablehead}
Method & Size (B) & Action P.T. & Spatial & Object & Goal & Long & Avg. \\
\midrule
\multicolumn{8}{@{}l}{\textcolor{gray}{\textit{\# Non-JEPA methods}}} \\
$\pi_{0.5}$~\cite{pi05} & 3.3 & \cmark & \underline{98.8} & 98.2 & 98.0 & 92.4 & 96.9 \\
\rowcolor{tablerowalt}
ABot-M0~\cite{yang2026abot} & 4.6 & \cmark & \underline{98.8} & \underline{99.8} & \textbf{99.0} & 96.6 & \underline{98.6} \\
ResVLA~\cite{resvla} & 4.0 & \xmark & 96.0 & \textbf{100.0} & 97.4 & 92.8 & 96.6 \\
\rowcolor{tablerowalt}
FastWAM~\cite{yuan2026fastwam} & 6.0 & \xmark & 98.2 & \textbf{100.0} & 97.0 & 95.2 & 97.6 \\
ImageWAM~\cite{imagewam} & 4.5 & \xmark & 97.2 & 99.2 & \underline{98.8} & \textbf{98.4} & 98.4 \\
\rowcolor{tablerowalt}
ForeWAM~\cite{huang2026foresight} & 1.5 & \xmark & 97.6 & \underline{99.8} & 98.0 & 96.2 & 97.9 \\
VLANeXt (Base)~\cite{vlanext} & 2.5 & \xmark & \textbf{99.0} & 99.2 & 96.6 & 94.8 & 97.4 \\
\midrule
\multicolumn{8}{@{}l}{\textcolor{gray}{\textit{\# JEPA-based methods}}} \\
\rowcolor{tablerowalt}
VLA-JEPA~\cite{vlajepa} & 2.3 & \cmark & 96.2 & 99.6 & 97.2 & 95.8 & 97.2 \\
JEPA-WAM~\cite{lin2026jepawam} & 1.3 & \xmark & 95.6 & 99.4 & 97.2 & 94.6 & 96.7 \\
\rowcolor{tablerowalt}
V-JEPA Policy (from scratch)~\cite{zhang2026vjepapolicy} & 0.9 & \xmark & 97.2 & 98.0 & 97.0 & 96.8 & 97.3 \\
\midrule
\rowcolor{thinkwamrow}
\thinkwam & 1.5 & \xmark & 98.4 & \textbf{100.0} & 98.4 & \underline{98.0} & \textbf{98.7} \\
\bottomrule
\end{tabularx}
\end{table*}

\begin{table*}[!t]
\centering
\caption{LIBERO-Plus success (\%) by perturbation category. For every method, Avg.\ is the unweighted mean of the seven categories, recomputed from the listed values.}
\label{tab:libero-plus}
\small
\renewcommand{\arraystretch}{0.90}
\setlength{\aboverulesep}{0.5pt}
\setlength{\belowrulesep}{0.75pt}
\setlength{\tabcolsep}{3pt}
\begin{tabularx}{\textwidth}{l*{8}{>{\raggedleft\arraybackslash}X}}
\toprule
\rowcolor{tablehead}
Method & Camera & Init. & Language & Light & Texture & Noise & Layout & Avg. \\
\midrule
\multicolumn{9}{@{}l}{\textcolor{gray}{\textit{\# Non-JEPA methods}}} \\
$\pi_{0.5}$~\cite{pi05} & 64.0 & 58.0 & \underline{88.5} & 96.6 & 81.4 & 87.5 & \textbf{85.9} & 80.3 \\
\rowcolor{tablerowalt}
ABot-M0~\cite{yang2026abot} & 60.4 & 67.9 & 86.4 & 96.2 & 91.6 & 86.4 & 82.6 & 81.6 \\
ResVLA~\cite{resvla} & 53.2 & 57.5 & 88.2 & 94.5 & \textbf{96.3} & 81.8 & 78.3 & 78.5 \\
\rowcolor{tablerowalt}
FastWAM~\cite{yuan2026fastwam} & 16.4 & 44.5 & 68.9 & 78.2 & 53.7 & 37.7 & 60.7 & 51.4 \\
ImageWAM~\cite{imagewam} & 80.8 & 50.3 & \textbf{91.4} & \textbf{98.1} & 85.5 & 93.8 & 80.5 & 82.9 \\
\rowcolor{tablerowalt}
ForeWAM~\cite{huang2026foresight} & 61.7 & 56.8 & 85.1 & 95.8 & 76.5 & 89.1 & 83.0 & 78.3 \\
VLANeXt (Base)~\cite{vlanext} & \underline{90.4} & 65.7 & 81.8 & 95.9 & 82.5 & \underline{94.1} & 80.8 & \underline{84.5} \\
\midrule
\multicolumn{9}{@{}l}{\textcolor{gray}{\textit{\# JEPA-based methods}}} \\
\rowcolor{tablerowalt}
VLA-JEPA~\cite{vlajepa} & 63.3 & 67.1 & 85.4 & 95.6 & 93.6 & 66.3 & \underline{85.1} & 79.5 \\
JEPA-WAM~\cite{lin2026jepawam} & 79.2 & 59.2 & 68.2 & 93.3 & \underline{94.6} & 83.6 & 76.1 & 79.2 \\
\rowcolor{tablerowalt}
V-JEPA Policy (from scratch)~\cite{zhang2026vjepapolicy} & 70.0 & \textbf{80.7} & 64.9 & 97.2 & 80.0 & 87.8 & 79.1 & 80.0 \\
\midrule
\rowcolor{thinkwamrow}
\thinkwam & \textbf{93.62} & \underline{74.19} & 74.82 & \underline{97.37} & 74.91 & \textbf{95.94} & 82.49 & \textbf{84.76} \\
\bottomrule
\end{tabularx}
\end{table*}

\begin{table}[!tbp]
\centering
\caption{RoboTwin 2.0 success (\%) on 20 fixed tasks. P.T.: action pretraining; ``---'': conventional policy sizes omitted.}
\label{tab:robotwin}
\footnotesize
\renewcommand{\arraystretch}{0.90}
\setlength{\aboverulesep}{0.5pt}
\setlength{\belowrulesep}{0.75pt}
\setlength{\tabcolsep}{3pt}
\begin{tabularx}{\linewidth}{Xccccc}
\toprule
\rowcolor{tablehead}
Method & Size (B) & P.T. & Clean $\uparrow$ & Random $\uparrow$ & Avg. $\uparrow$ \\
\midrule
Diffusion Policy~\cite{chi2023diffusionpolicy} & --- & \xmark & 48.0 & 1.6 & 24.8 \\
\rowcolor{tablerowalt}
ACT~\cite{zhao2023learningfinegrainedbimanualmanipulation} & --- & \xmark & 51.8 & 4.0 & 27.9 \\
RDT-1B~\cite{liu2024rdt1b} & 1.2 & \cmark & 56.0 & 24.1 & 40.1 \\
\rowcolor{tablerowalt}
DP3~\cite{ze2024dp3} & --- & \xmark & 73.9 & 8.3 & 41.1 \\
$\pi_0$~\cite{black2024pi0} & 3.3 & \cmark & 62.5 & 23.9 & 43.2 \\
\rowcolor{tablerowalt}
GO-1~\cite{bi2025motus} & 2.82 & \cmark & 56.75 & \underline{55.60} & 56.2 \\
$\pi_{0.5}$~\cite{pi05} & 3.3 & \cmark & 75.4 & 37.2 & 56.3 \\
\rowcolor{tablerowalt}
JEPA-WAM~\cite{lin2026jepawam} & 1.3 & \xmark & 79.9 & 36.9 & 58.4 \\
$\pi_{0.5}$ + JEPA Obj.~\cite{lin2026jepawam} & 3.3 & \cmark & 84.6 & 37.5 & 61.1 \\
\rowcolor{tablerowalt}
PAVE~\cite{pave2026} & 1.3 & \xmark & \textbf{86.8} & 45.2 & \underline{66.0} \\
\midrule
\rowcolor{thinkwamrow}
\thinkwam & 1.5 & \xmark & \underline{85.75} & \textbf{65.85} & \textbf{75.8} \\
\bottomrule
\end{tabularx}
\end{table}

\newsavebox{\thinkwamRoboCasaMainTable}
\begin{lrbox}{\thinkwamRoboCasaMainTable}
\begin{minipage}{\columnwidth}
\captionsetup{type=table}
\centering
\caption{RoboCasa24 average success (\%). Demos denotes training demonstrations per task.}
\label{tab:robocasa24}
\small
\renewcommand{\arraystretch}{1.00}
\setlength{\aboverulesep}{0.5pt}
\setlength{\belowrulesep}{0.75pt}
\setlength{\tabcolsep}{3pt}
\begin{tabularx}{\linewidth}{l*{3}{>{\centering\arraybackslash}X}c}
\toprule
\rowcolor{tablehead}
Method & Size (B) & P.T. & Demos & Avg. $\uparrow$ \\
\midrule
UWM~\cite{zheng2025flare} & 0.4 & \cmark & 1{,}000 & 60.8 \\
\rowcolor{tablerowalt}
$\pi_0$~\cite{zhang2026devadecoupledvideoactionmodel} & 3.3 & \cmark & 300 & 62.5 \\
GR00T-N1.5~\cite{zhang2026devadecoupledvideoactionmodel} & 3.0 & \cmark & 300 & 64.1 \\
\rowcolor{tablerowalt}
GR00T-N1~\cite{zhang2026dig} & 2.2 & \cmark & 50 & 36.0 \\
$\pi_{0.5}$~\cite{zhang2026dig} & 3.3 & \cmark & 50 & 41.4 \\
\rowcolor{tablerowalt}
UVA~\cite{liang2025videopolicy} & 0.5 & \xmark & 50 & 50.0 \\
Video Policy~\cite{liang2025videopolicy} & 2.43 & \xmark & 50 & 63.0 \\
\rowcolor{tablerowalt}
FastWAM~\cite{huang2026foresight} & 6.0 & \xmark & 50 & 49.5 \\
ForeWAM~\cite{huang2026foresight} & 1.5 & \xmark & 50 & 59.2 \\
\midrule
\rowcolor{thinkwamrow}
\thinkwam & 1.5 & \xmark & 50 & \textbf{65.17} \\
\bottomrule
\end{tabularx}
\end{minipage}
\end{lrbox}

\begin{table}[!tbp]
\centering
\usebox{\thinkwamRoboCasaMainTable}
\end{table}

\begin{table}[!tbp]
\centering
\caption{A1Z success (\%; $30$ trials per model/condition) and mean latency (ms/chunk). P.T.: action pretraining.}
\label{tab:realrobot_results}
\footnotesize
\renewcommand{\arraystretch}{1.10}
\setlength{\aboverulesep}{0.5pt}
\setlength{\belowrulesep}{0.75pt}
\setlength{\tabcolsep}{2pt}
\begin{tabularx}{\columnwidth}{lc*{3}{>{\centering\arraybackslash}X}cc}
\toprule
\rowcolor{tablehead}
Method & P.T. & \shortstack{Two\\targets} & \shortstack{$+$ Color\\distractors} & \shortstack{$+$ Purple\\light} & \shortstack{Mean\\success $\uparrow$} & \shortstack{Mean\\latency $\downarrow$} \\
\midrule
$\pi_{0.5}$ & \cmark & 66.7 & 50.0 & 53.3 & 56.7 & 119.9 \\
\rowcolor{tablerowalt}
FastWAM & \xmark & 46.7 & 30.0 & 36.7 & 37.8 & 296.9 \\
\midrule
\rowcolor{thinkwamrow}
\thinkwam & \xmark & \textbf{73.3} & \textbf{56.7} & \textbf{63.3} & \textbf{64.4} & \textbf{101.2} \\
\bottomrule
\end{tabularx}
\end{table}

\subsection{Experimental Setup}
\paragraph{Benchmarks and data.}
Our simulation evaluation covers LIBERO~\cite{liu2023libero}, LIBERO-Plus~\cite{fei2025libero-plus}, RoboTwin 2.0~\cite{chen2025robotwin}, and RoboCasa24~\cite{nasiriany2024robocasalargescalesimulationeveryday}. We train jointly on the four LIBERO suites and evaluate the same policy on LIBERO-Plus without OOD fine-tuning. On RoboTwin 2.0, we train on Clean demonstrations and evaluate the 20 tasks of JEPA-WAM Appendix D under both the Clean and Random settings. On RoboCasa24, we use 50 demonstrations per task and evaluate 50 rollouts per task. Real-world pick-and-place tests assess generalization from single-object training to two targets, with distractor or illumination perturbations (Sec.~\ref{sec:realrobot}).

\paragraph{Training and evaluation.}
For simulation, we follow each benchmark's protocol and report task success rate (\%). LIBERO-Plus averages are recomputed as unweighted means over the seven categories for all methods. Among the RoboTwin baselines, GO-1 is trained on mixed Clean and Random demonstrations and PAVE uses additional environment interaction. Implementation and evaluation details are in Appendices~\ref{app:config}--\ref{app:evaluation}.

\begin{figure}[!tbp]
\centering
\captionsetup{width=\columnwidth}
\includegraphics[width=\columnwidth,trim=3.5bp 0bp 0bp 0bp,clip]{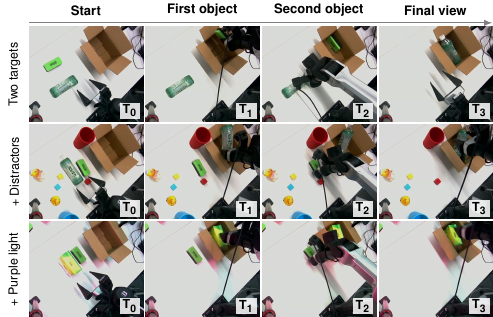}
\caption{ThinkWAM rollouts on the A1Z: two targets, color distractors, and purple light (top to bottom). Each row shows four chronological frames from a fixed third-person view.}
\label{fig:realrobot}
\end{figure}

\subsection{Simulation Results}
\paragraph{LIBERO.}
ThinkWAM reaches the highest average success, 98.70\% (Table~\ref{tab:libero}), improving over the JEPA-based methods by 1.4 to 2.0 percentage points, with the largest gains on Spatial and Long.

\paragraph{LIBERO-Plus.}
ThinkWAM achieves the highest macro-averaged success rate across the seven categories, 84.76\% (Table~\ref{tab:libero-plus}), and the best Camera and Noise robustness.

\begingroup
\paragraph{RoboTwin 2.0.}
\leavevmode\clubpenalty=0\relax
ThinkWAM leads with a 75.8\% average (Table~\ref{tab:robotwin}) and exceeds PAVE by 20.65 percentage points \mbox{on Random.}
\par\endgroup

\paragraph{RoboCasa24.}
ThinkWAM achieves the highest average success, 65.17\%, without action pretraining (Table~\ref{tab:robocasa24}). It outperforms Video Policy and ForeWAM with the same 50 demonstrations per task.

\begin{table}[!t]
\centering
\caption{Ablations on LIBERO and LIBERO-Plus (success, \%). R: proposal--rethink--refine; A: affordance guidance. $^{\dagger}$Variant 2 retains both passes but uses base futures in the second pass.}
\label{tab:component-ablation}
\label{tab:no-rethinking75k}
\footnotesize
\setlength{\tabcolsep}{3pt}
\renewcommand{\cmark}{\textcolor{tablecheck}{\scalebox{0.9}{\Checkmark}}}
\renewcommand{\xmark}{\textcolor{tablecross}{\scalebox{0.9}{\XSolidBrush}}}
\renewcommand{\arraystretch}{0.9}
\setlength{\aboverulesep}{1pt}
\setlength{\belowrulesep}{1.5pt}
\begin{tabularx}{\linewidth}{cXcccc}
\toprule
\rowcolor{tablehead}
ID & Configuration & R & A & LIBERO & LIBERO-Plus \\
\midrule
\multicolumn{6}{@{}l}{\textcolor{gray}{\textit{\# Multi-horizon future targets}}} \\
\rowcolor{thinkwamrow}
-- & \textbf{Full ThinkWAM} & \cmark & \cmark & \textbf{98.70} & \textbf{84.76} \\
1 & w/o rethink & \xmark & \cmark & 96.35 & 80.51 \\
\rowcolor{tablerowalt}
2\rlap{$^\dagger$} & w/o latent correction & \xmark & \cmark & 96.55 & 81.59 \\
3 & w/o affordance & \cmark & \xmark & 97.10 & 82.01 \\
\rowcolor{tablerowalt}
4 & w/o both & \xmark & \xmark & 95.00 & 76.41 \\
\midrule
\multicolumn{6}{@{}l}{\textcolor{gray}{\textit{\# Single future target}}} \\
\rowcolor{tablerowalt}
5 & single future & \cmark & \cmark & 96.40 & 79.92 \\
6 & \quad w/o rethink & \xmark & \cmark & 95.65 & 77.36 \\
\rowcolor{tablerowalt}
7 & \quad w/o affordance & \cmark & \xmark & 96.10 & 78.66 \\
8 & \quad w/o both & \xmark & \xmark & 94.30 & 75.58 \\
\bottomrule
\end{tabularx}
\end{table}

\begin{figure}[!b]
\centering
\captionsetup{width=\columnwidth}
\includegraphics[width=\columnwidth]{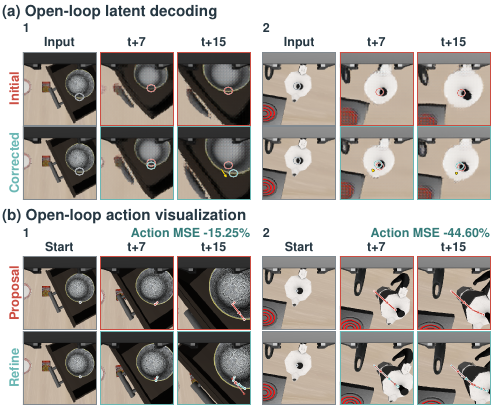}
\caption{Two paired open-loop examples at $h=0,7,15$, with $h=0$ the observed input. (a) Base and residual-corrected futures decoded by a shared frozen decoder; circles and arrows mark changes. (b) Proposal (red) and refined (teal) executions; dashed lines retain proposal paths and arrows indicate time.}
\label{fig:prediction-refinement}
\end{figure}

\begin{figure*}[!t]
\centering
\captionsetup{width=\textwidth}
\includegraphics[width=\textwidth]{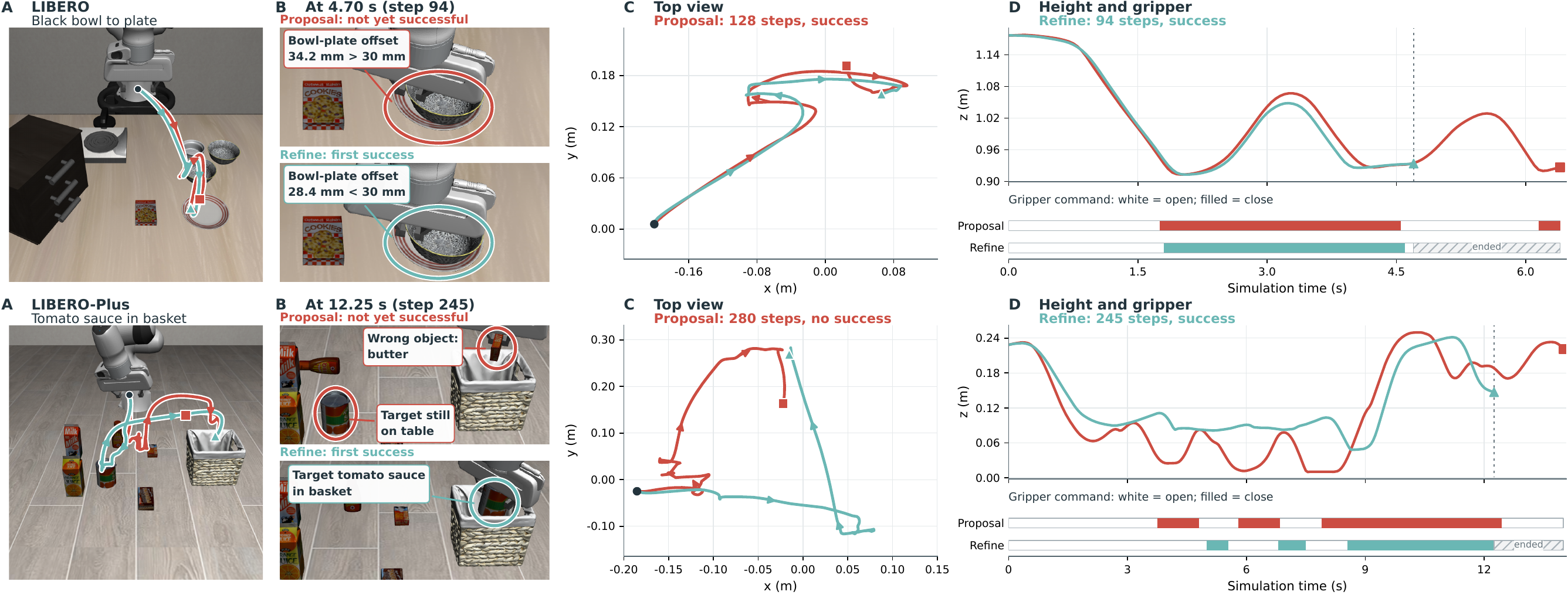}
\caption{Closed-loop comparison of proposal (red) and refined (teal) actions from matched initial states on LIBERO (top) and LIBERO-Plus (bottom); each branch replans from its own observations. (A) Scene trajectories. (B) Snapshots at the refined branch's first success. (C) Horizontal end-effector paths; bottom-row curves are slightly offset near their shared start for visibility. (D) End-effector height and gripper commands. Time excludes inference latency.}
\label{fig:closed-loop-action-comparison}
\end{figure*}

\subsection{Real-World Evaluation}
\label{sec:realrobot}
On the Galaxea A1Z, training data contain only demonstrations of picking up one green object and placing it into a nearby box. Training and testing use the same prompt: ``Pick up every green object and place it into the nearby box.'' We test three OOD settings without OOD fine-tuning: two green targets, added color distractors, and purple light; success requires both targets in the box. With $30$ trials per method per condition, \thinkwam{} achieves $64.4\%$ mean success, versus $56.7\%$ for $\pi_{0.5}$ and $37.8\%$ for FastWAM (Table~\ref{tab:realrobot_results}). Figure~\ref{fig:realrobot} shows rollouts.

\subsection{Ablation Study}
\label{sec:ablation}
Table~\ref{tab:component-ablation} compares inference-time interventions on rethinking, latent correction, and affordance guidance under multi-horizon and single-target prediction. Variants within each setting share one checkpoint (Appendix~\ref{app:ablation-protocol}).

Executing the proposal lowers success on both benchmarks, including a decrease from 84.76\% to 80.51\% on LIBERO-Plus. Retaining both action passes without latent correction reaches 81.59\%, compared with 84.76\% for the full model. This gap supports the contribution of proposal-conditioned future correction beyond using a second action pass. Rethinking yields a smaller gain with a single future target (variants 5--6). Affordance guidance adds complementary gains, while disabling both rethinking and affordance gives the lowest success in either setting.

\subsection{Inference Efficiency}
\label{sec:efficiency}
\begin{table}[!tbp]
\centering
\caption{Mean inference latency per action chunk (ms) on one A100 80GB.}
\label{tab:latency}
\small
\renewcommand{\arraystretch}{1.15}
\setlength{\aboverulesep}{0.5pt}
\setlength{\belowrulesep}{0.75pt}
\setlength{\tabcolsep}{3pt}
\begin{tabularx}{\linewidth}{>{\hsize=1.30\hsize\linewidth=\hsize\raggedright\arraybackslash}X>{\hsize=0.85\hsize\linewidth=\hsize\centering\arraybackslash}X>{\hsize=0.85\hsize\linewidth=\hsize\centering\arraybackslash}X}
\toprule
\rowcolor{tablehead}
Method & LIBERO $\downarrow$ & RoboTwin 2.0 $\downarrow$ \\
\midrule
LaWAM~\cite{lawam2026} & 222.73 & 215.62 \\
\rowcolor{tablerowalt}
$\pi_{0.5}$~\cite{pi05} (RLinf SFT~\cite{rlinf2026repository}) & 286.62 & 254.59 \\
OpenWAM-$\alpha$~\cite{wang2026openwam} & 413.97 & 426.83 \\
\rowcolor{tablerowalt}
FastWAM~\cite{yuan2026fastwam} & 506.48 & 496.16 \\
\midrule
\rowcolor{thinkwamrow}
\thinkwam & \textbf{153.48} & \textbf{164.61} \\
\bottomrule
\end{tabularx}
\end{table}

Including visual encoding and both action passes, \thinkwam{} takes 153.48\,ms per action chunk on LIBERO, 164.61\,ms on RoboTwin 2.0, and 101.2\,ms on real-world tasks (Table~\ref{tab:latency} and Table~\ref{tab:realrobot_results}). In simulation, these latencies are below those of all measured baselines under their native chunk shapes and integration steps.

\subsection{Design Analysis}
\label{sec:analysis}

\paragraph{Why condition the futures on the proposal?}
As noted in Sec.~\ref{sec:imagination}, the base prediction is identical for every proposal, and a shared backbone alone does not make predicted futures and actions agree~\cite{cyclewam2026}; rethinking adds the missing path from the proposal to the futures. Figure~\ref{fig:prediction-refinement} connects open-loop action executions and decoded future frames from the same examples at matching times. In (a), decoding the base and residual-corrected futures with the same frozen decoder shows how visual residual correction adjusts predicted object locations at both future horizons. In the two paired examples in (b), the refined chunks have 15.25\% and 44.60\% lower action MSE than the proposals, respectively.

\paragraph{What changes in manipulation?}
Figure~\ref{fig:closed-loop-action-comparison} shows two selected closed-loop comparisons of proposal and refined actions. From matched initial states, both branches run the same two-pass model at each replan: one executes the proposal $A_t^{(0)}$ and the other the chunk $A_t^{(1)}$ regenerated from the proposal-updated futures, with each branch continuing from its own observations. In LIBERO, the refined branch completes the bowl placement earlier, while the proposal branch is still misaligned with the plate at that moment. In LIBERO-Plus, the proposal branch transports a non-target object and fails within the step budget, whereas the refined branch places the target object and succeeds.

\section{Conclusion and Limitations}
\label{sec:conclusion}
\label{sec:limitations}
We presented \thinkwam{}, a latent world-action model that rethinks its actions with latent future feedback. It establishes bidirectional feedback before execution: the policy's own action proposal updates the multi-horizon future latents through a learned residual, and these updated futures then condition a second pass of the same action head, which generates the executable chunk. Affordance-guided memory and a near-fine, far-coarse token allocation supply the context and future representations that support this process. \thinkwam{} achieves high success rates and robustness on LIBERO, LIBERO-Plus, RoboTwin 2.0, RoboCasa24, and real-world manipulation tasks at low inference latency.

\thinkwam{} uses one round of proposal feedback before execution, requiring two action passes per chunk. Whether further iterations justify their extra cost remains untested; splitting one denoising trajectory between the two passes (\eg, 70\%/30\%) could reduce this cost.

\paragraph{Acknowledgments.}
We thank the Phantrix Research team for providing computational resources and technical support.
{
    \small
    \makeatletter
    \@secpenalty=0
    \renewcommand\bibsection{\@startsection{section}{1}{\z@}{-6pt plus -1pt minus -1pt}{4pt}{\large\bf}*{\refname}}%
    \makeatother
    \bibliographystyle{ieeenat_fullname}
    \bibliography{main}
}

\let\paragraph\thinkwamSavedParagraph
\appendix
\phantomsection
\pdfbookmark[0]{Appendix}{thinkwam-appendix}
\label{page:supp-start}
\raggedbottom
\renewcommand{\arraystretch}{0.90}
\setlength{\tabcolsep}{3pt}
\setlength{\aboverulesep}{0.5pt}
\setlength{\belowrulesep}{0.75pt}
\captionsetup[table]{skip=3pt}
\setlength{\intextsep}{6pt plus 2pt minus 2pt}
\setlength{\textfloatsep}{8pt plus 2pt minus 2pt}
\setlength{\dbltextfloatsep}{8pt plus 2pt minus 2pt}
\setlength{\floatsep}{6pt plus 2pt minus 2pt}
\setlength{\dblfloatsep}{6pt plus 2pt minus 2pt}
\setcounter{topnumber}{5}
\setcounter{bottomnumber}{5}
\setcounter{totalnumber}{10}
\setcounter{dbltopnumber}{5}
\renewcommand{\bottomfraction}{0.95}
\renewcommand{\topfraction}{0.95}
\renewcommand{\dbltopfraction}{0.95}
\renewcommand{\textfraction}{0.05}
\renewcommand{\dblfloatpagefraction}{0.95}

\makeatletter
\renewcommand\paragraph{\@startsection{paragraph}{4}{\z@}%
  {6pt plus 1pt minus 1pt}{-1em}{\normalfont\normalsize\bfseries}}
\makeatother

\section{Implementation, Training, and Inference}
\label{app:config}

\subsection{Model Configuration and Parameter Counts}
\label{app:model-config}
\paragraph{Pretrained components.}
The visual encoder is the released V-JEPA 2.1 ViT-L checkpoint~\cite{murlabadia2026vjepa21}. The language backbone is Qwen2.5-0.5B~\cite{qwen2025qwen25}, with learned projections that supply visual tokens as its input prefix.

Table~\ref{tab:app-config} records the architecture and benchmark adapters. The RoboCasa24 column describes the evaluated 40k-step P1 checkpoint; its replan interval is 16 control steps.

\begin{table*}[!tbp]
\centering
\caption{ThinkWAM architecture and benchmark-specific configuration.}
\label{tab:app-config}
\small
\begin{tabularx}{\textwidth}{@{}>{\columncolor{white}[0pt][\tabcolsep]\raggedright\arraybackslash}p{0.28\textwidth}*{2}{>{\raggedright\arraybackslash}X}>{\columncolor{white}[\tabcolsep][0pt]\raggedright\arraybackslash}X@{}}
\toprule
\rowcolor{tablehead}
Component & LIBERO & RoboTwin 2.0 & RoboCasa24 \\
\midrule
Model size & 1.507B & 1.501B & 1.505B \\
\rowcolor{tablerowalt}
External action pretraining & None & None & None \\
Replan & 14 & 32 & 16 \\
\rowcolor{tablerowalt}
Action chunk & 16 steps $\times$ 7 dims & 32 steps $\times$ 14 dims & 16 steps $\times$ 7 dims \\
Future horizons & $t+7$, $t+15$, $t+31$ & $t+8$, $t+16$, $t+32$ & $t+7$, $t+15$, $t+31$ \\
\rowcolor{tablerowalt}
Query token counts & 288 / 144 / 72 & 288 / 144 / 72 & 288 / 144 / 72 \\
Memory slots & 16 & 16 & 16 \\
\rowcolor{tablerowalt}
Action-history length & 8 & 16 & 8 \\
Cameras & 2 & 3 & 3 \\
\rowcolor{tablerowalt}
Integration steps (1st / 2nd pass) & 4 / 4 (Euler) & 4 / 4 (Euler) & 4 / 4 (Euler) \\
Training proposal steps $S_{\mathrm{train}}$ & 1 & 1 & 1 \\
\rowcolor{tablerowalt}
$\lambda_0$ / $\lambda_1$ & 1.2 / 1.4 & 1.2 / 1.4 & 1.0 / 1.0 \\
$\lambda_{\mathrm{jepa}}$ / $\lambda_{\mathrm{res}}$ & 0.5 / 0.2 & 0.5 / 0.2 & 0.5 / 0.2 \\
\rowcolor{tablerowalt}
$\lambda_{\mathrm{Dice}}$ / $\lambda_{\mathrm{aff}}$ (max) & 0.5 / 0.05 & -- / 0 & 0.5 / 0.1 \\
Affordance guidance & Enabled & Disabled & Enabled \\
\rowcolor{tablerowalt}
Affordance labels & Contact heatmaps, $24\times24$ soft labels & -- & Same as LIBERO \\
Memory-fusion weight $\bar\alpha$ (max) & 0.05 & -- & 0.05 \\
\rowcolor{tablerowalt}
Latent residual & \multicolumn{3}{>{\columncolor{white}[\tabcolsep][0pt]}l@{}}{Action-conditioned, three horizons} \\
\bottomrule
\end{tabularx}
\end{table*}

The 1.507B LIBERO model comprises V-JEPA (304.7M), Qwen (513.4M), ActionDiT (536.6M), and the remaining 152.1M for the temporal predictor, memory, latent residual predictor, affordance decoder, and adapters.

\paragraph{Module specifications and parameter sharing.}
Table~\ref{tab:app-module-specs} lists the module dimensions of the reported checkpoints. The complete ActionDiT is shared between the proposal and refinement passes.

\begin{table}[!tbp]
\centering
\caption{Module specifications. FFN is the intermediate feed-forward dimension. The proposal context has 80 tokens (64 action placeholders and 16 memory slots); refinement appends 504 corrected future tokens.}
\label{tab:app-module-specs}
\small
\setlength{\tabcolsep}{3pt}
\begin{tabularx}{\columnwidth}{@{}>{\columncolor{white}[0pt][\tabcolsep]\raggedright\arraybackslash}Xrrr>{\columncolor{white}[\tabcolsep][0pt]}r@{}}
\toprule
\rowcolor{tablehead}
Module & Width & Blocks & Heads & FFN \\
\midrule
V-JEPA 2.1 ViT-L/16 (frozen) & 1,024 & 24 & 16 & 4,096 \\
\rowcolor{tablerowalt}
Qwen2.5-0.5B & 896 & 24 & 14 & 4,864 \\
Memory & 1,024 & 2 & 8 & 4,096 \\
\rowcolor{tablerowalt}
Temporal and latent-residual predictors & 1,024 & 2 & 8 & 4,096 \\
Affordance decoder & 256 & 2 & 8 & 1,024 \\
\rowcolor{tablerowalt}
Shared ActionDiT & 1,536 & 16 & 32 & 6,144 \\
\bottomrule
\end{tabularx}
\end{table}

\paragraph{Input and action interfaces.}
Each RGB view is resized to $384\times384$, normalized with ImageNet statistics, encoded separately, and concatenated in a fixed camera order (LIBERO: agent, wrist; RoboTwin: high, left wrist, right wrist; RoboCasa24: left agent, right agent, wrist). Instructions are tokenized by Qwen and padded or truncated to 128 tokens. LIBERO uses the stored 7D controller commands and an 8D end-effector and gripper state; RoboTwin standardizes its 14D actions and states with dataset statistics; RoboCasa24 maps actions to $[-1,1]$ with the released low/high statistics and uses a zero proprioception placeholder.

\subsection{Multi-Horizon Targets and Latent Residuals}
\label{app:multihorizon-implementation}

\paragraph{Temporal windows and target construction.}
At time $t$, the history contains only previously executed actions, $a_{t-L},\ldots,a_{t-1}$; it is right-aligned and left-padded with zeros near the episode start. The action target is the complete chunk beginning at $t$, and each visual target is taken from the recorded observation at its configured offset $t+h$. Padded action positions are masked in both flow-matching losses; per-horizon validity masks are supplied to the JEPA and latent-residual objectives.

\begin{figure}[!htb]
\centering
\includegraphics[width=\linewidth,trim=3bp 0bp 21bp 0bp,clip]{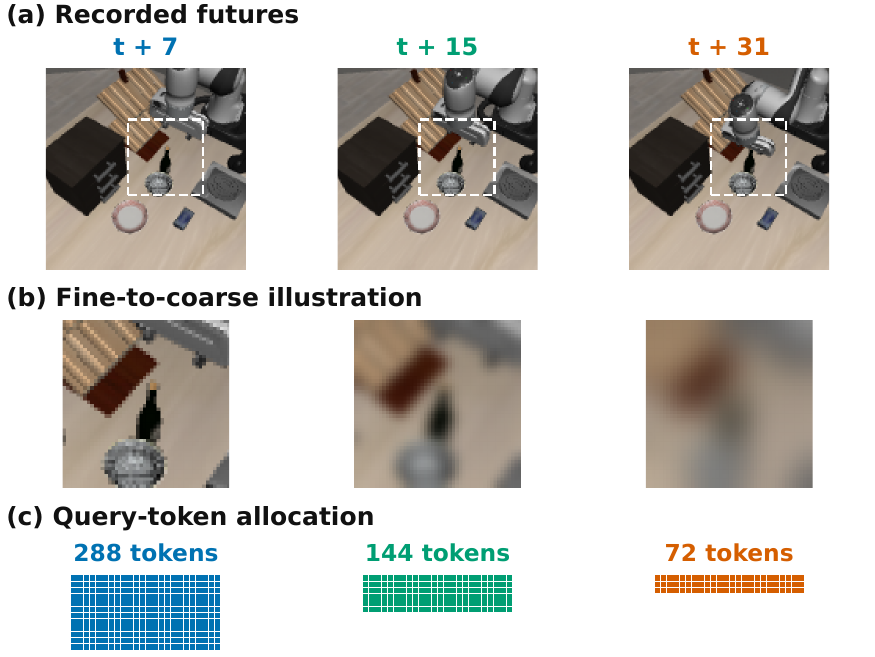}
\caption{Multi-horizon targets and token budgets. (a) Recorded observations at $t+7$, $t+15$, and $t+31$. (b) One crop at increasing blur, illustrating fine-to-coarse detail. (c) The 288/144/72 query tokens, one cell per token.}
\label{fig:horizon-tokens75k}
\end{figure}

\paragraph{Latent scales.}
JEPA supervision uses L2-normalized pooled future features, whereas residual regression and latent addition operate on unnormalized features; the corrected latents are L2-normalized before projection into the action-conditioning context. Figure~\ref{fig:horizon-tokens75k} illustrates the 288/144/72 token allocation.

\paragraph{What the latent residual encodes.}
The base prediction $\hat z_{t+h}$, supervised by demonstrated futures, provides a visual reference for action correction. Its scene and task information is useful but not necessarily in the form that best supports correcting a specific proposal. Given the proposal $A_t^{(0)}$, the residual model therefore learns an incremental adjustment, and the refinement path reads the base prediction through a stop-gradient, $\tilde z_{t+h}=\operatorname{sg}[\hat z_{t+h}]+R_h(z_t,A_t^{(0)})$. The forward pass still uses $\hat z_{t+h}$, but through this addition the second-pass action loss reaches only the residual and does not directly modify $\hat z_{t+h}$, which remains trained by $\mathcal{L}_{\mathrm{jepa}}$. Since $r_h^*$ is defined by the demonstrated future rather than by the outcome of executing $A_t^{(0)}$, \thinkwam{} does not claim to predict the consequences of the proposal. Two signals determine the adjustment. First, $\mathcal{L}_{\mathrm{res}}$, applied with a small weight ($\lambda_{\mathrm{res}}=0.2$) in the high-dimensional V-JEPA space, keeps it visually anchored, guiding the corrected latents toward the expert future in overall direction. Second, $\lambda_1\mathcal{L}_{\mathrm{FM}}(c_t^{(1)},A^*)$ back-propagates through $\Delta z_{t+h}$ and trains the residual to bridge the gap between the base future prediction and the action proposal, mapping $A_t^{(0)}$ onto the re-corrected visual representation that conditions the second pass. The residual thus performs visually constrained, proposal-conditioned representation adaptation: as the visual representation is corrected, it becomes better fitted to the second action pass and reflects the mapping of the proposed actions.

Supervising the residual with the actual outcomes of executing $A_t^{(0)}$ would require on-policy rollouts during training, which substantially slows training in simulation and is impractical on real robots. We therefore use demonstrated futures as the supervision signal and leave outcome-supervised variants to future work.

\subsection{Affordance Supervision and Memory Fusion}
\label{app:affordance-implementation}
The affordance labels are contact heatmaps. The heatmaps are normalized by their per-frame, per-view maximum and area-averaged to $24\times24$ per view without binarization. The loss is BCE on logits against these soft labels plus $\lambda_{\mathrm{Dice}}$ times a smoothed Dice loss, masked for samples without labels and without class weighting. In the affordance summary, the certainty weights $\rho_{t,n}$ are stop-gradient, and the output layer of $g_{\mathrm{aff}}$ is zero-initialized so that training starts from the unguided memory. The fusion weight is $\alpha_t=\bar\alpha(k)\,\gamma(\bar\rho_t)$, where $\bar\alpha(k)$ is the step schedule given below and $\gamma$ is a sigmoid gate on the mean certainty $\bar\rho_t$.

\paragraph{Guidance schedules.}
On LIBERO, $\lambda_{\mathrm{aff}}$ warms up to 0.05 over 3k steps, while $\bar\alpha$ and the upstream-gradient scale (which only rescales gradients into the shared features) warm up linearly from 0 to 0.05; the gate is $\gamma(\bar\rho)=\operatorname{sigmoid}((\bar\rho-0.70)/0.10)$. RoboCasa24 uses $\lambda_{\mathrm{aff}}=0.1$ with $\bar\alpha$ and the upstream-gradient scale fixed at 0.05, and RoboTwin disables the branch. Because $\Delta m_t^{\mathrm{aff}}$ is computed from $U_t$, which already contains $m_t$, the temporal context and future predictor are evaluated once per decision step and are not rerun after memory fusion.

\begin{figure*}[!t]
\centering
\includegraphics[width=\textwidth,trim=6bp 0bp 4.5bp 0bp,clip]{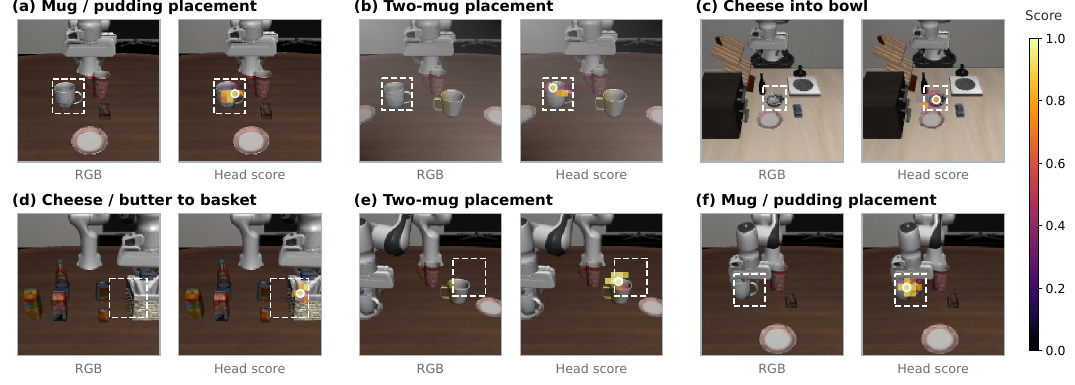}
\caption{Affordance-head outputs on LIBERO. Each example pairs an RGB observation with the predicted affordance scores on a shared 0--1 scale. Dashed boxes identify task objects, and circles locate score maxima.}
\label{fig:affordance75k}
\end{figure*}

\paragraph{Affordance visualizations.}
Figure~\ref{fig:affordance75k} shows affordance-head outputs on LIBERO.

\subsection{Optimization and Training Algorithm}
\label{app:training-algorithm}

\paragraph{Optimization and compute.}
All runs use AdamW ($\beta_1,\beta_2=0.9,0.95$, weight decay 0), BF16, and global gradient-norm clipping at 1.0, with linear warmup followed by cosine decay. LIBERO and RoboCasa24 P0 use a global batch of 96 (24 per GPU); RoboTwin and RoboCasa24 P1 use 80 (20 per GPU), without gradient accumulation. All runs use four A100-SXM4-80GB GPUs. GPU-hours in Table~\ref{tab:app-training-stages} are estimated from training logs. LIBERO trains 40k steps in P0 and 40k in P1.

\begin{table}[!htbp]
\centering
\caption{Training budgets and estimated GPU-hours.}
\label{tab:app-training-stages}
\small
\renewcommand{\arraystretch}{0.9}
\setlength{\tabcolsep}{3pt}
\begin{tabularx}{\linewidth}{@{}>{\columncolor{white}[0pt][\tabcolsep]\raggedright\arraybackslash}Xcc>{\columncolor{white}[\tabcolsep][0pt]}r@{}}
\toprule
\rowcolor{tablehead}
Stage & Steps & LR  & GPU-h \\
\midrule
LIBERO P0 & 40k & $10^{-4}\to10^{-5}$ & 75.4 \\
\rowcolor{tablerowalt}
LIBERO P1 & 40k & $10^{-4}\to10^{-5}$ & 116.5 \\
RoboTwin & 90k & $5\times10^{-5}\to5\times10^{-6}$ & 211.7 \\
\rowcolor{tablerowalt}
RoboCasa24 P0 & 20k & $10^{-4}\to10^{-5}$ & 43.0 \\
RoboCasa24 P1 & 40k & $10^{-4}\to10^{-6}$ & 88.0 \\
\bottomrule
\end{tabularx}
\end{table}

P0 optimizes one action flow-matching loss with coefficient 1 and the JEPA loss with coefficient 0.5. P1 introduces the latent residual, the second action loss, and affordance supervision where enabled; its loss coefficients are in Table~\ref{tab:app-config}. The visual encoder itself remains frozen, while the current-observation adapter provides the configured visual adaptation.

\paragraph{Training graph.}
Algorithm~\ref{alg:training} gives the optimization graph for P0 and P1. The frozen visual encoder defines the future targets throughout, and the observation adapter is applied only to the current features, before memory and language-action encoding.

For action supervision, $\mathcal{L}_{\mathrm{FM}}(c,A^*)$ regresses the flow velocity toward $A^*-\epsilon$ at $x_\tau=(1-\tau)\epsilon+\tau A^*$, with $\epsilon\sim\mathcal{N}(0,I)$. The implementation draws $b\sim\operatorname{Beta}(1.5,1.0)$ and sets $\tau=(0.999-b)/0.999$. The two action losses independently use the same expert chunk $A^*$. $\operatorname{Rollout}_{\theta}$ integrates the shared action head differentiably to obtain the proposal that conditions the residual model. Line 19 detaches only the path from $\hat z_{t+h}$ into $\tilde z_{t+h}$ (Appendix~\ref{app:multihorizon-implementation}).

\begin{algorithm}[!tbp]
\caption{ThinkWAM training: initialization and proposal-conditioned refinement}
\label{alg:training}
\footnotesize
\begin{algorithmic}[1]
\Require Demonstrations $\mathcal{D}$, stage $s\in\{\mathrm{P0},\mathrm{P1}\}$, frozen encoder $E_v$, horizons $\mathcal{H}$, budgets $\{N_h\}$
\State Initialize the configured trainable modules; load the configured warm-start or continuation checkpoint when applicable
\For{each training step $k$}
  \State Sample $(o_t,\ell,q_t,\mathcal{A}_t,A^*,\{o_{t+h}\},y^{\mathrm{aff}})$ from the training split
  \State Apply the configured learning-rate and guidance schedules
  \State $z_t\gets E_v(o_t)$; apply the current-observation adapter when enabled in P1
  \State $u_h\gets\operatorname{sg}[\operatorname{Pool}_{N_h}(E_v(o_{t+h}))]$, $y_h\gets\operatorname{Norm}(u_h)$ for every $h\in\mathcal{H}$
  \State $(H_t^v,f_t^{\mathrm{lang-act}})\gets\operatorname{Qwen}(z_t,\ell)$; $m_t\gets\mathcal{M}(z_t,\mathcal{A}_t)$
  \State $(\{\hat z_{t+h}\},U_t)\gets\operatorname{TemporalJEPA}(H_t^v,m_t)$
  \State $\mathcal{L}_{\mathrm{jepa}}\gets\operatorname{Mean}_{h}\!\left[\operatorname{Mean}_{i}[1-\cos(\hat z_{t+h,i},y_{h,i})]\right]$
  \If{$s=\mathrm{P0}$}
    \State $c_t\gets[f_t^{\mathrm{lang-act}};W_m m_t]$
    \State $\mathcal{L}\gets\lambda_{\mathrm{act}}\mathcal{L}_{\mathrm{FM}}(c_t,A^*)+\lambda_{\mathrm{jepa}}\mathcal{L}_{\mathrm{jepa}}$
  \Else
    \State Decode affordance from $U_t$'s visual prefix with the scheduled upstream-gradient scale
    \State $m_t^{\mathrm{guide}}\gets m_t+\bar\alpha(k)\,\gamma(\bar\rho_t)\,\Delta m_t^{\mathrm{aff}}$; use zero guidance when disabled
    \State $c_t^{(0)}\gets[f_t^{\mathrm{lang-act}};W_m m_t^{\mathrm{guide}}]$
    \State $A_t^{(0)}\gets\operatorname{Rollout}_{\theta}(c_t^{(0)},q_t,\epsilon_0;S_{\mathrm{train}})$, $\epsilon_0\sim\mathcal{N}(0,I)$
    \State $\Delta z_{t+h}\gets R_h(z_t,A_t^{(0)})$; $r_h^*\gets\operatorname{sg}[u_h-\hat z_{t+h}]$ for every $h\in\mathcal{H}$
    \State $\tilde z_{t+h}\gets\operatorname{sg}[\hat z_{t+h}]+\Delta z_{t+h}$ for every $h\in\mathcal{H}$
    \State $c_t^{(1)}\gets[c_t^{(0)};\operatorname{Concat}_h W_h\operatorname{Norm}(\tilde z_{t+h})]$
    \State $\mathcal{L}_{\mathrm{res}}\gets\operatorname{Mean}_h\operatorname{SmoothL1}(\Delta z_{t+h},r_h^*)$
    \State $\mathcal{L}_{\mathrm{aff}}\gets\mathrm{BCE}+\lambda_{\mathrm{Dice}}\mathrm{Dice}$ on available affordance labels, or zero when disabled
    \State $\mathcal{L}\gets\lambda_0\mathcal{L}_{\mathrm{FM}}(c_t^{(0)},A^*)+\lambda_1\mathcal{L}_{\mathrm{FM}}(c_t^{(1)},A^*)$
    \State \hspace{1.5em}$+\lambda_{\mathrm{jepa}}\mathcal{L}_{\mathrm{jepa}}+\lambda_{\mathrm{res}}\mathcal{L}_{\mathrm{res}}+\lambda_{\mathrm{aff}}(k)\mathcal{L}_{\mathrm{aff}}$
  \EndIf
  \State Backpropagate $\mathcal{L}$, clip the gradient norm, and update the active parameters with AdamW
  \State Save checkpoints at fixed intervals
\EndFor
\end{algorithmic}
\end{algorithm}

\subsection{Inference Procedure}
\label{app:inference}
Deployment follows Sec.~\ref{sec:objective}. Both passes use four Euler integration steps, whereas the differentiable training proposal uses $S_{\mathrm{train}}=1$. For chunk length $K$ and replanning interval $R$, the policy executes the first $R$ actions of $A_t^{(1)}$, appends them to the action history, and replans from the next observation (LIBERO: $K=16$, $R=14$; RoboTwin: $K=32$, $R=32$; RoboCasa24: $K=16$, $R=16$).

\section{Experimental Protocols}
\label{app:evaluation}

\subsection{Benchmark Data and Evaluation Settings}
\label{app:evaluation-settings}
\paragraph{LIBERO.}
The fixed evaluation covers all 40 tasks in the Spatial, Object, Goal, and Long suites, using the first 50 initial states per task (2,000 records in total). Native task predicates determine success. The same checkpoint is used for LIBERO and LIBERO-Plus, with no parameter updates during evaluation. Per-task results appear in Appendix~\ref{app:libero-tasks}.

\paragraph{LIBERO-Plus.}
Evaluation uses all 10,030 registry variants at initial state 0. Category denominators differ because the released manifest contains different numbers of valid records per category. LIBERO-Plus category records are Camera 1{,}599, Initial state 1{,}550, Language 1{,}537, Light 1{,}142, Texture 1{,}076, Noise 1{,}601 and Layout 1{,}525 (10{,}030 in total); the reported 84.76\% is the unweighted category mean, while the micro average over all records is 84.84\%. Suite-by-category counts are listed in Appendix~\ref{app:plus-breakdown}.

\paragraph{RoboTwin 2.0.}
\label{app:robotwin-protocol}
Training uses 50 tasks with 50 Clean demonstrations each (2,500 in total); evaluation uses the fixed 20-task subset from JEPA-WAM Appendix~D. Each task contributes 100 episodes under each of the Clean and Random conditions, giving 2,000 records per condition. The Clean/Random average is $(\mathrm{Clean}+\mathrm{Random})/2$. The per-task results in Appendix~\ref{app:robotwin-results} give 1,715/2,000 (85.75\%) for Clean and 1,317/2,000 (65.85\%) for Random. RoboTwin qualitative examples use the same checkpoint and are described in Appendix~\ref{app:robotwin-qualitative}.

\paragraph{RoboCasa24.}
We use a nominal budget of 50 demonstrations per task, with three camera views and a 7D action interface. The evaluation protocol uses 50 rollouts per task. Table~\ref{tab:app-config} records the configuration of the evaluated checkpoint.

\paragraph{Training-data organization.}
The LIBERO training set combines the standard demonstrations of all 40 tasks. LIBERO and RoboTwin use a 90/10 split of training windows (seed 7), and LIBERO training applies image augmentation with probability 0.5, disabled at validation. RoboCasa24 is split by episode into disjoint training and validation sets. Action-history padding and future-window validity are specified in Appendix~\ref{app:multihorizon-implementation}.

\subsection{Baseline Sources and Comparison Conventions}
\label{app:baseline-sources}

\paragraph{LIBERO and LIBERO-Plus.}
In Tables~\ref{tab:libero} and~\ref{tab:libero-plus}, seven baseline rows follow V-JEPA Policy~\cite{zhang2026vjepapolicy}, Table~1. The exceptions are ABot-M0~\cite{yang2026abot}, Tables~3/4 (zero-shot transfer from LIBERO to LIBERO-Plus); ForeWAM~\cite{huang2026foresight}, Tables~10/5; and VLANeXt (Base)~\cite{vlanext}, Tables~2/3, respectively. Row citations identify the original methods. LIBERO averages are retained as reported. Every LIBERO-Plus Avg.\ is recomputed as the unweighted mean of the seven listed categories, matching ThinkWAM.
\paragraph{RoboTwin sources and training protocols.}
In Table~\ref{tab:robotwin}, Diffusion Policy, ACT, RDT-1B, DP3, $\pi_0$, $\pi_{0.5}$, JEPA-WAM, and $\pi_{0.5}$+JEPA Obj.\ are transcribed from JEPA-WAM~\cite{lin2026jepawam}, Table~3 and Appendix~D. GO-1~\cite{agibot2025colosseo} uses the per-task Clean/Random results in Motus~\cite{bi2025motus} (v2, Table~14), re-averaged over these same 20 tasks, giving 56.75\%/55.60\%. Its training uses mixed Clean and Random demonstrations across 50 tasks. PAVE is from PAVE~\cite{pave2026} (v3, Table~3) and additionally uses Clean-environment interaction. These training protocols differ from Clean-demonstration-only training.

\paragraph{RoboCasa24 scores.}
In Table~\ref{tab:robocasa24}, row citations identify score sources. The $\pi_0$ and GR00T-N1.5 results follow DeVA~\cite{zhang2026devadecoupledvideoactionmodel}, Table~1; GR00T-N1 and $\pi_{0.5}$ follow DiG~\cite{zhang2026dig}, Table~2; UVA~\cite{li2025unifiedvideoaction} and Video Policy follow Video Policy~\cite{liang2025videopolicy}, Table~1; UWM follows the reproduced baseline in FLARE~\cite{zheng2025flare}, Table~1, with its 1,000-demonstration budget recorded in DeVA~\cite{zhang2026devadecoupledvideoactionmodel}, Table~1; FastWAM and ForeWAM follow ForeWAM~\cite{huang2026foresight}, Table~3. Training demonstrations per task remain explicit in the main table.
\paragraph{Model-size conventions.}
Model sizes are reported in billions of parameters. Video Policy's 2.43B is counted from its officially released RoboCasa checkpoint~\cite{liang2025videopolicy}, including frozen text and VAE modules; GR00T-N1 follows its original paper~\cite{nvidia2025groot}; UVA follows ViPRA~\cite{vipra2026videoactions}, Table~9; and ABot-M0 follows V-JEPA Policy~\cite{zhang2026vjepapolicy}, Table~2 (VLM and action expert, excluding optional spatial modules). V-JEPA Policy excludes its frozen text encoder, and sizes are omitted for the conventional Diffusion Policy, ACT, and DP3 baselines. Action P.T.\ (P.T.\ in Tables~\ref{tab:robotwin} and~\ref{tab:robocasa24}) denotes additional robot-policy pretraining with action labels, including inherited pretrained policy weights; visual, language, or video initialization alone does not count. For UWM, the reported model size and action-pretraining status refer to its original pretrained model.

\paragraph{RoboCasa24 action pretraining.}
In Table~\ref{tab:robocasa24}, $\pi_0$, $\pi_{0.5}$, GR00T-N1, and GR00T-N1.5 inherit action-pretrained policies. UVA is the Video Policy study's reproduced baseline; its visual initialization and Video Policy's video-only first stage do not count as action pretraining~\cite{liang2025videopolicy}. FastWAM and ForeWAM follow the no-policy-pretraining setting of the cited RoboCasa comparison~\cite{huang2026foresight}. ForeWAM's frozen LaWM teacher nevertheless carries external embodied pretraining, which its policy-level P.T.\ convention excludes.

\subsection{Latency Measurement Protocol}
\label{app:latency-protocol}

\begin{table}[!tbp]
\centering
\caption{Action chunk shapes. Shapes are chunk length $\times$ output dimension; parentheses indicate native or padded dimensions.}
\label{tab:app-latency-settings}
\small
\setlength{\tabcolsep}{3pt}
\begin{tabularx}{\columnwidth}{@{}>{\hsize=1.10\hsize\linewidth=\hsize\columncolor{white}[0pt][\tabcolsep]\raggedright\arraybackslash}X>{\hsize=0.80\hsize\linewidth=\hsize\centering\arraybackslash}X>{\hsize=1.10\hsize\linewidth=\hsize\columncolor{white}[\tabcolsep][0pt]\centering\arraybackslash}X@{}}
\toprule
\rowcolor{tablehead}
Method & LIBERO & RoboTwin \\
\midrule
LaWAM & $8\times32$ {\footnotesize (7 native)} & $36\times32$ {\footnotesize (16 native EEF)} \\
\addlinespace[1pt]
\rowcolor{tablerowalt}
$\pi_{0.5}$ {\footnotesize (RLinf SFT)} & $50\times7$ & $50\times14$ {\footnotesize (32 padded)} \\
\addlinespace[1pt]
OpenWAM-$\alpha$ & $32\times10$ {\footnotesize EEF} & $32\times20$ {\footnotesize EEF} \\
\addlinespace[1pt]
\rowcolor{tablerowalt}
FastWAM & $32\times7$ & $32\times14$ \\
\midrule
\rowcolor{thinkwamrow}
\thinkwam & $16\times7$ & $32\times14$ \\
\bottomrule
\end{tabularx}
\end{table}
Table~\ref{tab:latency} reports mean latencies over 60 timed calls after 10 warm-up calls on a single A100 80GB at batch size one, keeping all samples without outlier removal. Inputs are synthetic RGB and state with real task instructions; simulator rendering and model initialization are excluded. Each method retains its native action-block shape (Table~\ref{tab:app-latency-settings}). The four baselines use 10 integration steps; \thinkwam uses two passes with four Euler steps each. These values measure per-chunk generation cost; chunk length and replanning interval are separate quantities.

\subsection{Real-Robot Evaluation Protocol}
\label{app:realrobot}

\paragraph{Platform and training task.}
We compare ThinkWAM, FastWAM, and $\pi_{0.5}$ on the Galaxea A1Z single-arm platform. Training demonstrations show one green object being picked up and placed into a nearby box. The training instruction is ``Pick up every green object and place it into the nearby box.'' This instruction is used verbatim at test time.

\paragraph{Three OOD conditions.}
All test conditions contain two green targets, and both must be placed in the box. \textbf{Two targets} increases the number of green objects from one to two. \textbf{Two targets + color distractors} adds objects of other colors to this two-target scene; the two green objects remain the targets. \textbf{Two targets + purple light} applies purple illumination to the two-target scene, without adding the distractors of the second condition. Thus, all three conditions include a target-count shift; the latter two additionally test color-based target selection or robustness to illumination. None of these three conditions is an in-distribution single-object test.

\paragraph{Matched evaluation and success.}
We evaluate each method for 30 trials per condition, giving 90 trials per method and 270 trials across all three methods. Each trial is one episode requiring both placements, rather than two independently scored single-object trials. An episode succeeds only when both green targets are placed in the box within the common execution time limit; placing only one is a failure. Methods use matched initial placements, the same instruction, and identical success criteria. All use policies trained on the single-object demonstrations, with no additional fine-tuning on any of the three OOD conditions.

Table~\ref{tab:realrobot_results} reports complete-episode success rates. In the order of two targets, color distractors, and purple light, the success counts are $22/30$, $17/30$, and $19/30$ for ThinkWAM; $20/30$, $15/30$, and $16/30$ for $\pi_{0.5}$; and $14/30$, $9/30$, and $11/30$ for FastWAM.

\paragraph{Aggregation.}
For each method and condition, the success rate is $100s/30\%$, where $s$ is the number of successful episodes. With equal denominators, the mean over the three OOD conditions is $100(s_{\mathrm{two}}+s_{\mathrm{distr}}+s_{\mathrm{purple}})/90\%$. The totals are $58/90$ ($64.4\%$) for ThinkWAM, $51/90$ ($56.7\%$) for $\pi_{0.5}$, and $34/90$ ($37.8\%$) for FastWAM. The 270 trials cover all three methods and are not the denominator for ThinkWAM alone. Failed episodes, timeouts, and interventions are recorded under the same predefined rules.

\paragraph{Inference latency.}
Table~\ref{tab:realrobot_results} reports measured mean inference latency per action chunk on the same NVIDIA RTX 4080 SUPER with batch size $1$ and non-RTC execution. Each model is measured over $120$ calls after warm-up, separately from the $30$ task-success trials per OOD condition. ThinkWAM generates $32$ actions with four steps in each of its two stages. $\pi_{0.5}$ generates $50$ actions with $10$ steps. FastWAM generates $32$ actions with $10$ steps. These measurements use the stated model-specific configurations and chunk lengths; they are per-chunk inference times, not per-action latency or physical task-completion time.

\paragraph{Image provenance.}
Figure~\ref{fig:realrobot} shows recorded ThinkWAM executions in the two-target, color-distractor, and purple-light conditions. Each row uses four chronological frames from a single rollout, captured by the fixed third-person \texttt{head\_rgb} camera. The original wrist-camera view is excluded.

\section{Additional Ablations}
\label{app:ablations}
This section specifies the component ablations in Table~\ref{tab:component-ablation}.

\subsection{Ablation Evaluation Protocol}
\label{app:ablation-protocol}
\label{app:full-proposal75k}
\paragraph{Interventions.}
The eight configurations correspond to the IDs in Table~\ref{tab:component-ablation}. Variants 1--4 use the full-model checkpoint. Variant 1 executes the initial proposal with affordance guidance retained. Variant 2 keeps both action passes but feeds the second pass the base latents $\hat z_{t+h}$ instead of the corrected latents $\tilde z_{t+h}$, so the refined chunk is regenerated from futures that carry no proposal information; the first pass and the latent residual are unchanged. Variant 3 keeps the proposal--rethink--refine loop but zeros the affordance-to-memory residual at inference; it does not remove the head's training supervision retroactively. Variant 4 applies both switches. Variant 5 replaces the multi-horizon objective with a single $t+15$ target with 576 query tokens, which roughly matches the total token count of the three multi-horizon targets (288+144+72). Variants 6--8 apply the interventions of variants 1, 3, and 4, respectively, to the variant-5 checkpoint. Only variant 5 requires its own training run.

\paragraph{Evaluation sets and metrics.}
All variants use the evaluation sets and native success predicates in Appendix~\ref{app:evaluation-settings}. All LIBERO-Plus entries in the ablation table use the unweighted mean of the seven category success rates; LIBERO averages over the 2,000 states.

\subsection{Real-Robot Ablation: Failure Cases}
\label{app:realrobot-failures}
\begin{figure}[!tbp]
    \centering
    \includegraphics[width=\columnwidth]{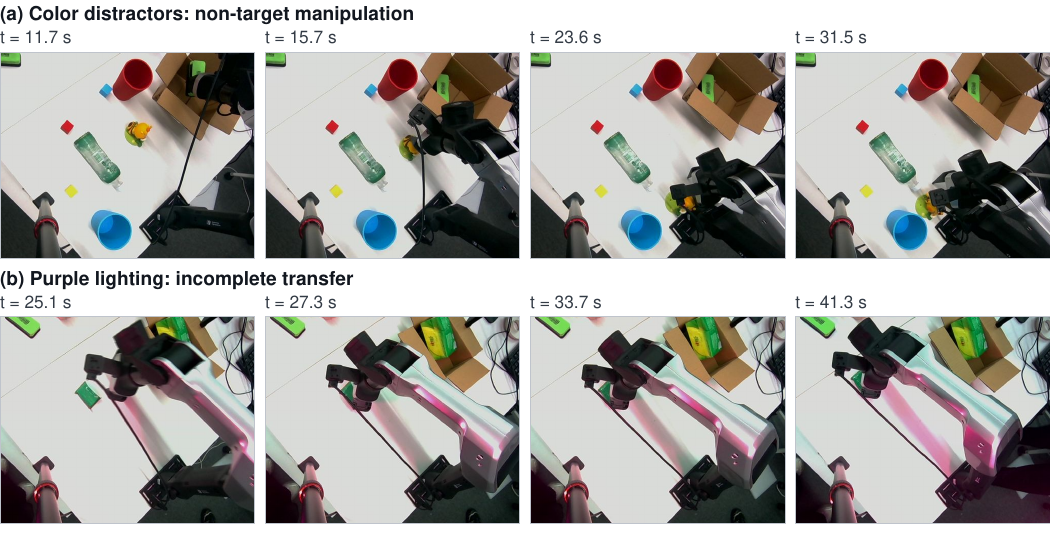}
    \caption{\textbf{Real-robot failure cases of ThinkWAM w/o rethink.} Each row shows four chronological third-person frames from one episode: (a) manipulation of a non-target object amid color distractors, and (b) an incomplete transfer under purple illumination.}
    \label{fig:realrobot-failures}
\end{figure}

We additionally examine real-robot executions of ThinkWAM \emph{w/o rethink}, which executes the initial action proposal without the rethink--refine stage. Figure~\ref{fig:realrobot-failures} presents two selected failure episodes on the Galaxea A1Z. In the distractor scene, the gripper manipulates a yellow non-target object while a green target remains on the table. Under purple illumination, the gripper approaches an object but does not complete its transfer into the box during the displayed interval. These examples document failure behavior of the real-robot ablation qualitatively. Adding rethinking improves success in simulation (Table~\ref{tab:component-ablation}) and may help reduce similar target-selection and incomplete-transfer errors on the real robot.

\section{Detailed Benchmark Results}
\label{app:detailed-results}
The following tables expand the main-paper benchmark results using the evaluation units and aggregation rules in Appendix~\ref{app:evaluation}.

\subsection{LIBERO Task-Level Results}
\label{app:libero-tasks}
Table~\ref{tab:app-libero-task} lists every LIBERO task in the final record set. The suite-level totals reproduce the main-paper score of 1,974 successes out of 2,000 records.

\begin{table}[!tbp]
\centering
\caption{Per-task LIBERO success (50 initial states per task, 2,000 records).}
\label{tab:app-libero-task}
\footnotesize
\renewcommand{\arraystretch}{0.90}
\setlength{\tabcolsep}{2.5pt}
\begin{tabularx}{\linewidth}{@{}>{\columncolor{white}[0pt][\tabcolsep]}l>{\raggedright\arraybackslash}X>{\columncolor{white}[\tabcolsep][0pt]}r@{}}
\toprule
\rowcolor{tablehead}
Task ID & Task instruction & SR (\%) \\
\midrule
Long-0 & put both the alphabet soup and the tomato sauce in the basket & 100.00 \\
\rowcolor{tablerowalt}
Long-1 & put both the cream cheese box and the butter in the basket & 100.00 \\
Long-2 & turn on the stove and put the moka pot on it & 98.00 \\
\rowcolor{tablerowalt}
Long-3 & put the black bowl in the bottom drawer of the cabinet and close it & 98.00 \\
Long-4 & put the white mug on the left plate and put the yellow and white mug on the right plate & 94.00 \\
\rowcolor{tablerowalt}
Long-5 & pick up the book and place it in the back compartment of the caddy & 96.00 \\
Long-6 & put the white mug on the plate and put the chocolate pudding to the right of the plate & 100.00 \\
\rowcolor{tablerowalt}
Long-7 & put both the alphabet soup and the cream cheese box in the basket & 100.00 \\
Long-8 & put both moka pots on the stove & 100.00 \\
\rowcolor{tablerowalt}
Long-9 & put the yellow and white mug in the microwave and close it & 94.00 \\
Goal-0 & open the middle drawer of the cabinet & 100.00 \\
\rowcolor{tablerowalt}
Goal-1 & put the bowl on the stove & 98.00 \\
Goal-2 & put the wine bottle on top of the cabinet & 100.00 \\
\rowcolor{tablerowalt}
Goal-3 & open the top drawer and put the bowl inside & 94.00 \\
Goal-4 & put the bowl on top of the cabinet & 98.00 \\
\rowcolor{tablerowalt}
Goal-5 & push the plate to the front of the stove & 100.00 \\
Goal-6 & put the cream cheese in the bowl & 100.00 \\
\rowcolor{tablerowalt}
Goal-7 & turn on the stove & 100.00 \\
Goal-8 & put the bowl on the plate & 98.00 \\
\rowcolor{tablerowalt}
Goal-9 & put the wine bottle on the rack & 96.00 \\
\bottomrule
\end{tabularx}
\end{table}

\subsection{LIBERO-Plus Category and Suite Breakdown}
\label{app:plus-breakdown}
Table~\ref{tab:app-plus-breakdown} gives the success counts for each suite and perturbation category. The seven category rates are averaged without weighting to obtain the reported 84.76\% macro average.

\subsection{RoboTwin Task-Level Results}
\label{app:robotwin-results}
Table~\ref{tab:app-robotwin-protocol} lists the fixed 20-task subset, with 100 episodes per task in each condition. Baseline sources and training-data differences are recorded in Appendix~\ref{app:baseline-sources}.

\subsection{RoboCasa24 Task-Level Results}
\label{app:robocasa-results}
Table~\ref{tab:app-robocasa-tasks} lists the per-task success rates of the final 40k-step P1 checkpoint, with an overall score of 65.17\% as reported in Table~\ref{tab:robocasa24}.

\begin{table}[!tbp]
\ContinuedFloat
\centering
\caption[]{Per-task LIBERO success (continued).}
\label{tab:app-libero-task-continued}
\footnotesize
\renewcommand{\arraystretch}{0.90}
\setlength{\tabcolsep}{2.5pt}
\begin{tabularx}{\linewidth}{@{}>{\columncolor{white}[0pt][\tabcolsep]}l>{\raggedright\arraybackslash}X>{\columncolor{white}[\tabcolsep][0pt]}r@{}}
\toprule
\rowcolor{tablehead}
Task ID & Task instruction & SR (\%) \\
\midrule
Object-0 & pick up the alphabet soup and place it in the basket & 100.00 \\
\rowcolor{tablerowalt}
Object-1 & pick up the cream cheese and place it in the basket & 100.00 \\
Object-2 & pick up the salad dressing and place it in the basket & 100.00 \\
\rowcolor{tablerowalt}
Object-3 & pick up the bbq sauce and place it in the basket & 100.00 \\
Object-4 & pick up the ketchup and place it in the basket & 100.00 \\
\rowcolor{tablerowalt}
Object-5 & pick up the tomato sauce and place it in the basket & 100.00 \\
Object-6 & pick up the butter and place it in the basket & 100.00 \\
\rowcolor{tablerowalt}
Object-7 & pick up the milk and place it in the basket & 100.00 \\
Object-8 & pick up the chocolate pudding and place it in the basket & 100.00 \\
\rowcolor{tablerowalt}
Object-9 & pick up the orange juice and place it in the basket & 100.00 \\
Spatial-0 & pick up the black bowl between the plate and the ramekin and place it on the plate & 94.00 \\
\rowcolor{tablerowalt}
Spatial-1 & pick up the black bowl next to the ramekin and place it on the plate & 98.00 \\
Spatial-2 & pick up the black bowl from table center and place it on the plate & 100.00 \\
\rowcolor{tablerowalt}
Spatial-3 & pick up the black bowl on the cookie box and place it on the plate & 100.00 \\
Spatial-4 & pick up the black bowl in the top drawer of the wooden cabinet and place it on the plate & 100.00 \\
\rowcolor{tablerowalt}
Spatial-5 & pick up the black bowl on the ramekin and place it on the plate & 96.00 \\
Spatial-6 & pick up the black bowl next to the cookie box and place it on the plate & 100.00 \\
\rowcolor{tablerowalt}
Spatial-7 & pick up the black bowl on the stove and place it on the plate & 100.00 \\
Spatial-8 & pick up the black bowl next to the plate and place it on the plate & 98.00 \\
\rowcolor{tablerowalt}
Spatial-9 & pick up the black bowl on the wooden cabinet and place it on the plate & 98.00 \\
\bottomrule
\end{tabularx}
\end{table}

\begin{table}[!tbp]
\centering
\caption{LIBERO-Plus success counts by suite and perturbation category.}
\label{tab:app-plus-breakdown}
\small
\renewcommand{\arraystretch}{0.90}
\setlength{\tabcolsep}{3pt}
\begin{tabularx}{\linewidth}{@{}>{\columncolor{white}[0pt][\tabcolsep]}l>{\raggedright\arraybackslash}Xc>{\columncolor{white}[\tabcolsep][0pt]}r@{}}
\toprule
\rowcolor{tablehead}
Suite & Category & \mbox{Successes / total} & Rate (\%) \\
\midrule
Long & Camera Viewpoints & 367 / 419 & 87.59 \\
\rowcolor{tablerowalt}
Long & Robot Initial States & 277 / 393 & 70.48 \\
Long & Language Instructions & 335 / 383 & 87.47 \\
\rowcolor{tablerowalt}
Long & Light Conditions & 260 / 274 & 94.89 \\
Long & Background Textures & 143 / 289 & 49.48 \\
\rowcolor{tablerowalt}
Long & Sensor Noise & 417 / 449 & 92.87 \\
Long & Objects Layout & 276 / 312 & 88.46 \\
\rowcolor{tablerowalt}
Spatial & Camera Viewpoints & 361 / 376 & 96.01 \\
Spatial & Robot Initial States & 272 / 350 & 77.71 \\
\rowcolor{tablerowalt}
Spatial & Language Instructions & 319 / 390 & 81.79 \\
Spatial & Light Conditions & 288 / 292 & 98.63 \\
\rowcolor{tablerowalt}
Spatial & Background Textures & 239 / 258 & 92.64 \\
Spatial & Sensor Noise & 347 / 351 & 98.86 \\
\rowcolor{tablerowalt}
Spatial & Objects Layout & 354 / 385 & 91.95 \\
Object & Camera Viewpoints & 395 / 396 & 99.75 \\
\rowcolor{tablerowalt}
Object & Robot Initial States & 300 / 398 & 75.38 \\
Object & Language Instructions & 312 / 354 & 88.14 \\
\rowcolor{tablerowalt}
Object & Light Conditions & 296 / 297 & 99.66 \\
Object & Background Textures & 207 / 248 & 83.47 \\
\rowcolor{tablerowalt}
Object & Sensor Noise & 420 / 422 & 99.53 \\
Object & Objects Layout & 349 / 403 & 86.60 \\
\rowcolor{tablerowalt}
Goal & Camera Viewpoints & 374 / 408 & 91.67 \\
Goal & Robot Initial States & 301 / 409 & 73.59 \\
\rowcolor{tablerowalt}
Goal & Language Instructions & 184 / 410 & 44.88 \\
Goal & Light Conditions & 268 / 279 & 96.06 \\
\rowcolor{tablerowalt}
Goal & Background Textures & 217 / 281 & 77.22 \\
Goal & Sensor Noise & 352 / 379 & 92.88 \\
\rowcolor{tablerowalt}
Goal & Objects Layout & 279 / 425 & 65.65 \\
\bottomrule
\end{tabularx}
\end{table}

\begin{table}[!tbp]
\centering
\caption{Per-task ThinkWAM success on the fixed RoboTwin 20-task subset (100 episodes per task and setting).}
\label{tab:app-robotwin-protocol}
\small
\renewcommand{\arraystretch}{1.05}
\setlength{\tabcolsep}{3pt}
\begin{tabularx}{\linewidth}{@{}>{\columncolor{white}[0pt][\tabcolsep]\raggedright\arraybackslash}Xr>{\columncolor{white}[\tabcolsep][0pt]}r@{}}
\toprule
\rowcolor{tablehead}
Task & Clean (\%) & Random (\%) \\
\midrule
Adjust Bottle & 100.00 & 95.00 \\
\rowcolor{tablerowalt}
Beat Block Hammer & 52.00 & 24.00 \\
Click Alarmclock & 95.00 & 76.00 \\
\rowcolor{tablerowalt}
Click Bell & 100.00 & 88.00 \\
Dump Bin Bigbin & 97.00 & 85.00 \\
\rowcolor{tablerowalt}
Grab Roller & 99.00 & 78.00 \\
Handover Mic & 100.00 & 92.00 \\
\rowcolor{tablerowalt}
Lift Pot & 92.00 & 81.00 \\
Place Bread Basket & 58.00 & 39.00 \\
\rowcolor{tablerowalt}
Place Bread Skillet & 74.00 & 24.00 \\
Place Burger Fries & 98.00 & 63.00 \\
\rowcolor{tablerowalt}
Place Cans Plasticbox & 57.00 & 47.00 \\
Place Empty Cup & 77.00 & 61.00 \\
\rowcolor{tablerowalt}
Place Object Basket & 78.00 & 34.00 \\
Place Shoe & 84.00 & 56.00 \\
\rowcolor{tablerowalt}
Press Stapler & 89.00 & 61.00 \\
Shake Bottle Horizontally & 100.00 & 96.00 \\
\rowcolor{tablerowalt}
Shake Bottle & 100.00 & 82.00 \\
Stack Bowls Three & 72.00 & 52.00 \\
\rowcolor{tablerowalt}
Stack Bowls Two & 93.00 & 83.00 \\
\midrule
\rowcolor{thinkwamrow}
\textbf{Average} & \textbf{85.75} & \textbf{65.85} \\
\bottomrule
\end{tabularx}
\end{table}

\begin{table}[!tbp]
\centering
\caption{Per-task ThinkWAM success on RoboCasa24 (24 tasks).}
\label{tab:app-robocasa-tasks}
\footnotesize
\setlength{\tabcolsep}{2pt}
\begin{tabularx}{\linewidth}{Xr@{\hspace{10pt}}Xr}
\toprule
\rowcolor{tablehead}
Task & SR (\%) & Task & SR (\%) \\
\midrule
PnP\allowbreak Counter\allowbreak To\allowbreak Cab & 34.00 & Open\allowbreak Drawer & 86.00 \\
\rowcolor{tablerowalt}
PnP\allowbreak Cab\allowbreak To\allowbreak Counter & 32.00 & Close\allowbreak Drawer & 98.00 \\
PnP\allowbreak Counter\allowbreak To\allowbreak Sink & 26.00 & Turn\allowbreak On\allowbreak Stove & 70.00 \\
\rowcolor{tablerowalt}
PnP\allowbreak Sink\allowbreak To\allowbreak Counter & 40.00 & Turn\allowbreak Off\allowbreak Stove & 44.00 \\
PnP\allowbreak Counter\allowbreak To\allowbreak Microwave & 26.00 & Turn\allowbreak On\allowbreak Sink\allowbreak Faucet & 80.00 \\
\rowcolor{tablerowalt}
PnP\allowbreak Microwave\allowbreak To\allowbreak Counter & 34.00 & Turn\allowbreak Off\allowbreak Sink\allowbreak Faucet & 86.00 \\
PnP\allowbreak Counter\allowbreak To\allowbreak Stove & 48.00 & Turn\allowbreak Sink\allowbreak Spout & 76.00 \\
\rowcolor{tablerowalt}
PnP\allowbreak Stove\allowbreak To\allowbreak Counter & 48.00 & Coffee\allowbreak Setup\allowbreak Mug & 28.00 \\
Open\allowbreak Single\allowbreak Door & 84.00 & Coffee\allowbreak Serve\allowbreak Mug & 74.00 \\
\rowcolor{tablerowalt}
Close\allowbreak Single\allowbreak Door & 84.00 & Coffee\allowbreak Press\allowbreak Button & 90.00 \\
Open\allowbreak Double\allowbreak Door & 94.00 & Turn\allowbreak On\allowbreak Microwave & 90.00 \\
\rowcolor{tablerowalt}
Close\allowbreak Double\allowbreak Door & 94.00 & Turn\allowbreak Off\allowbreak Microwave & 98.00 \\
\midrule
\rowcolor{thinkwamrow}
\multicolumn{3}{l}{\textbf{Overall}} & \textbf{65.17} \\
\bottomrule
\end{tabularx}
\end{table}

\section{Qualitative Rollout Comparisons}
\label{app:additional}
\label{app:qualitative}

\subsection{Matched-Start Rollout Comparisons}
\label{app:paired-rollouts}
The following cases compare proposal-only and refined policies under matched initial conditions. Each benchmark uses its evaluated checkpoint, with replanning settings given in Table~\ref{tab:app-config}.

\begin{figure*}[!t]
\centering
\captionsetup{width=0.84\textwidth}
\includegraphics[width=0.84\textwidth]{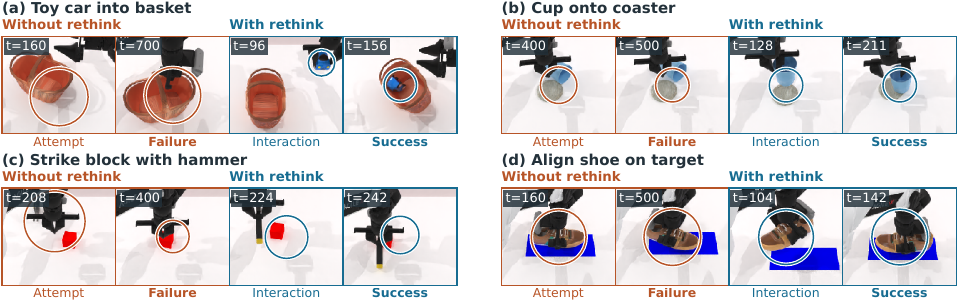}
\caption{RoboTwin rollouts from matched starts. For each task, the left pair shows the proposal-only interaction and the right pair the interaction and final success with rethink. Circles mark the interaction region.}
\label{fig:robotwin-recovery90k}
\end{figure*}
\begin{figure*}[!t]
\centering
\captionsetup{skip=3pt,width=\textwidth}
\includegraphics[width=0.84\textwidth]{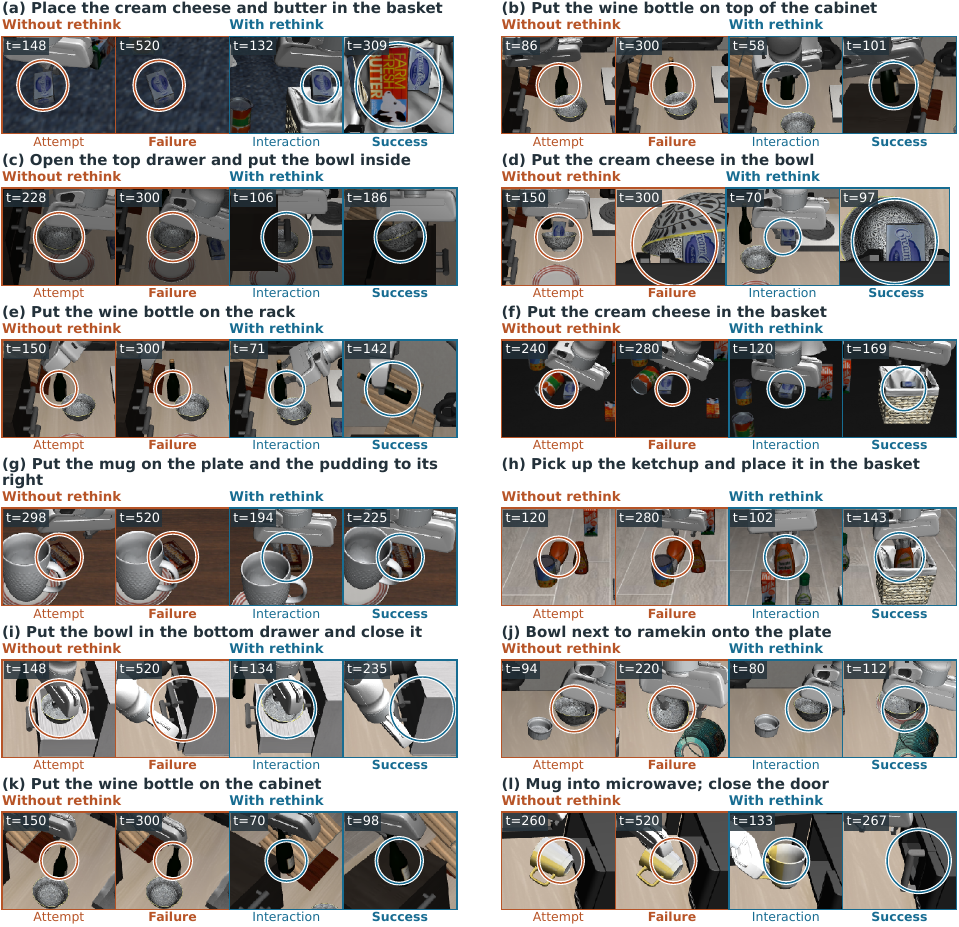}
\caption{LIBERO-family rollouts from matched starts. Each card contrasts the proposal-only interaction with the successful interaction with rethink. Circles mark the object region; timestamps belong to independent trajectories.}
\label{fig:recovery-cases1}
\label{fig:recovery-cases2}
\end{figure*}

\begin{figure*}[!t]
\centering
\includegraphics[width=\textwidth]{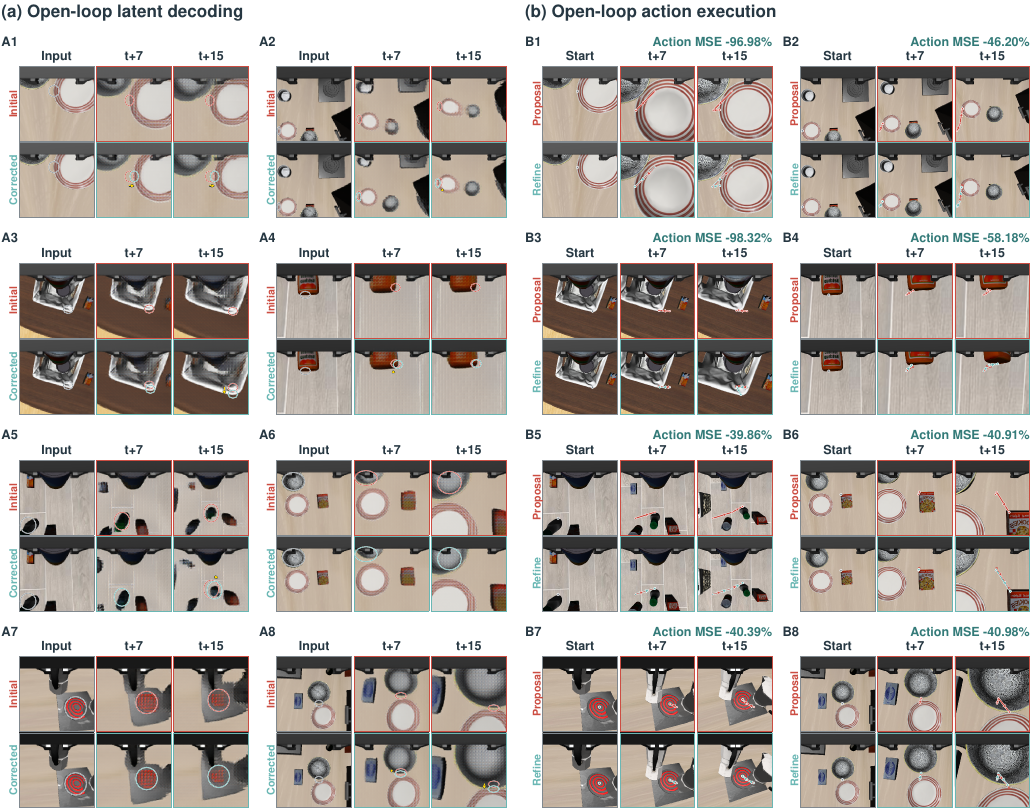}
\caption{Additional paired open-loop comparisons, following the visual conventions of Figure~\ref{fig:prediction-refinement}. A1--A8 and B1--B8 show decoded futures and action executions for the same examples at $h=0,7,15$ ($h=0$: observed input). For B1--B8, action MSE decreases by 96.98\%, 46.20\%, 98.32\%, 58.18\%, 39.86\%, 40.91\%, 40.39\%, and 40.98\%, respectively.}
\label{fig:appendix-openloop-comparisons}
\end{figure*}

\begin{figure*}[!t]
\centering
\captionsetup{skip=4pt}
\includegraphics[width=\textwidth]{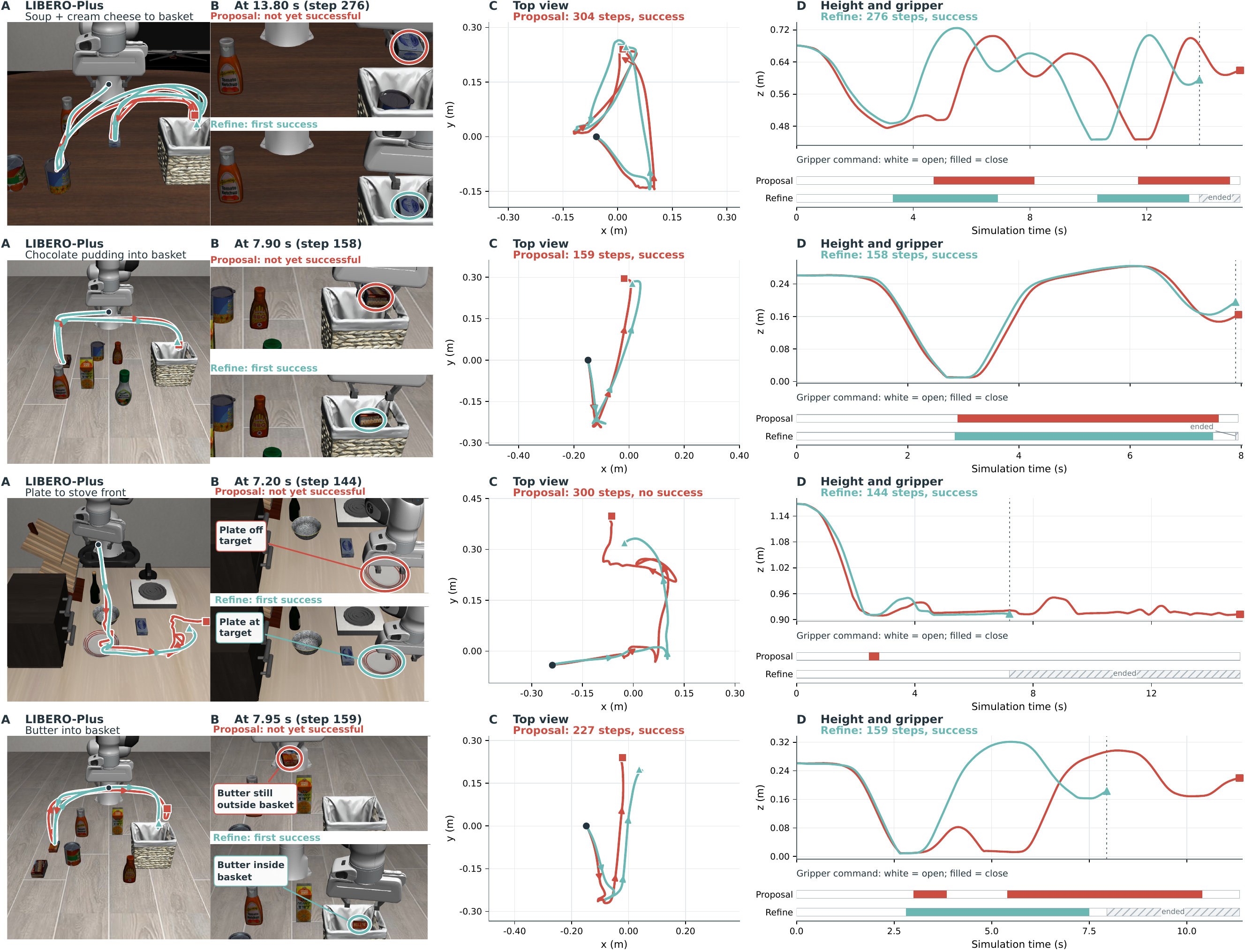}
\caption{Additional closed-loop comparisons following the protocol and layout of Figure~\ref{fig:closed-loop-action-comparison}; proposal in red, refinement in teal. \textbf{(A)} End-effector paths on the initial scene. \textbf{(B)} Snapshots at refinement's first success, when the proposal has not yet succeeded; circles and labels mark task-relevant differences. \textbf{(C)} Horizontal end-effector paths; black dots mark starts, squares and triangles mark proposal and refined endpoints. \textbf{(D)} End-effector height and gripper commands (white: open; filled: close). Rollouts stop at first success or the task budget (hatched: execution ended). Time is at 20\,Hz after initialization, excluding inference latency.}
\label{fig:appendix-closed-loop-comparisons}
\end{figure*}
\subsubsection{RoboTwin Recovery Trajectories}
\label{app:robotwin-qualitative}
Figure~\ref{fig:robotwin-recovery90k} shows RoboTwin rollouts in which rethinking corrects the contact interaction and completes the task. Both branches share the initial three-camera observations, scene seeds, instructions, and policy seeds.


\subsubsection{LIBERO Recovery Trajectories}
\label{app:recovery}
Figure~\ref{fig:recovery-cases1} shows LIBERO-family rollouts with matched inputs and policy seeds, in which the refined policy completes grasping, transport, placement, and drawer-closing tasks that the proposal alone does not.
\par\vspace{0pt}

\subsection{Additional Open-Loop Visualizations}
\label{app:open-loop-visualizations}
Figure~\ref{fig:appendix-openloop-comparisons} extends Figure~\ref{fig:prediction-refinement} with eight paired examples.

Each pair shares its demonstration, decision index, wrist-camera view, and policy input, including the configured expert history. One two-pass call produces both action chunks and latent versions, using independent sampling noise for the two action passes. Executions start from matched physical states and identically initialized controllers; frames show $h=0,7,15$ control steps at 20\,Hz.

For raw sampled actions, we compute $E^{(i)}=(16\cdot7)^{-1}\sum_{k=0}^{15}\|A_{t,k}^{(i)}-a^{\mathrm{expert}}_{t+k}\|_2^2$ and report $100(E^{(0)}-E^{(1)})/E^{(0)}$. Observations follow their same-index expert actions, so execution starts from the next stored state. All ten examples---the two in the main paper and the eight shown here---improve whole-action and arm-only MSE under both the training-indexed and one-step-shifted references, for raw samples and commands after clipping and gripper binarization.

Action paths connect manually marked image landmarks and include wrist-camera motion; B6 tracks a static Cookies-box corner. The same frozen decoder visualizes base and proposal-conditioned corrected latents.

\subsection{Additional Closed-Loop Comparisons}
\label{app:additional-closed-loop}
Figure~\ref{fig:appendix-closed-loop-comparisons} extends Figure~\ref{fig:closed-loop-action-comparison} with further LIBERO-Plus examples. The two branches in each pair start from identical simulator states and initial observations, then execute proposal and refined actions, respectively, while each replans from its own observations ($K=16$, $R=14$). In each case the refined branch reaches the goal earlier, with more accurate placement or the correct target object.

\end{document}